%% file: main.tex
\documentclass[sigconf]{acmart}
\AtBeginDocument{%
  \providecommand\BibTeX{{%
    \normalfont B\kern-0.5em{\scshape i\kern-0.25em b}\kern-0.8em\TeX}}}

\setcopyright{acmcopyright}
\copyrightyear{2018}
\acmYear{2018}
\acmDOI{10.1145/1122445.1122456}

\acmConference[Woodstock '18]{Woodstock '18: ACM Symposium on Neural
  Gaze Detection}{June 03--05, 2018}{Woodstock, NY}
\acmBooktitle{Woodstock '18: ACM Symposium on Neural Gaze Detection,
  June 03--05, 2018, Woodstock, NY}
\acmPrice{15.00}
\acmISBN{978-1-4503-XXXX-X/18/06}

\usepackage{algorithm}
\usepackage{algorithmic}
\input{math_command.tex}

\usepackage{multirow}
\usepackage{latexsym}
\usepackage{url}
\usepackage{amsmath,nccmath}
\usepackage{xcolor}
\usepackage{graphicx}
\usepackage{caption}
\usepackage{enumitem}
\usepackage{array}

\usepackage{subfigure}
\newcolumntype{P}[1]{>{\centering\arraybackslash}p{#1}}

\newcommand{\suhang}[1]{{\color{blue}[SW: #1]}}

\newcommand{\method}{CAIL}

\copyrightyear{2024}
\acmYear{2024}
\setcopyright{acmlicensed}\acmConference[WSDM '24]{Proceedings of the 17th ACM International Conference on Web Search and Data Mining}{March 4--8, 2024}{Merida, Mexico}
\acmBooktitle{Proceedings of the 17th ACM International Conference on Web Search and Data Mining (WSDM '24), March 4--8, 2024, Merida, Mexico}
\acmPrice{15.00}
\acmDOI{10.1145/3616855.3635827}
\acmISBN{979-8-4007-0371-3/24/03}

\begin{document}

\title{Interpretable Imitation Learning with Dynamic Causal Relations}

\author{Tianxiang Zhao}
\affiliation{%
  \institution{The Pennsylvania State University}
  \state{State College}
  \country{USA}
}
\email{tkz5084@psu.edu}

\author{Wenchao Yu}
\affiliation{%
  \institution{NEC Laboratories America}
  \state{Princeton}
  \country{USA}
}
\email{wyu@nec-labs.com}

\author{Suhang Wang}
\affiliation{%
  \institution{The Pennsylvania State University}
  \state{State College}
  \country{USA}
}
\email{szw494@psu.edu}

\author{Lu Wang}
\affiliation{%
  \institution{East China Normal University}
  \state{Shanghai}
  \country{China}
}
\email{luwang@stu.ecnu.edu.cn}

\author{Xiang Zhang}
\affiliation{%
  \institution{The Pennsylvania State University}
  \state{State College}
  \country{USA}
}
\email{xzz89@psu.edu}

\author{Yuncong Chen}
\affiliation{%
  \institution{NEC Laboratories America}
  \state{Princeton}
  \country{USA}
}
\email{yuncong@nec-labs.com}

\author{Yanchi Liu}
\affiliation{%
  \institution{NEC Laboratories America}
  \state{Princeton}
  \country{USA}
}
\email{yanchi@nec-labs.com}

\author{Wei Cheng}
\affiliation{%
  \institution{NEC Laboratories America}
  \state{Princeton}
  \country{USA}
}
\email{weicheng@nec-labs.com}

\author{Haifeng Chen}
\affiliation{%
  \institution{NEC Laboratories America}
  \state{Princeton}
  \country{USA}
}
\email{haifeng@nec-labs.com}

\renewcommand{\shortauthors}{Tianxiang Zhao et al.}

\begin{abstract}
Imitation learning, which learns agent policy by mimicking expert demonstration, has shown promising results in many applications such as medical treatment regimes and self-driving vehicles. However, it remains a difficult task to interpret control policies learned by the agent. Difficulties mainly come from two aspects: 1) agents in imitation learning are usually implemented as deep neural networks, which are black-box models and lack interpretability; 2) the latent causal mechanism behind agents' decisions may vary along the trajectory, rather than staying static throughout time steps. To increase transparency and offer better interpretability of the neural agent, we propose to expose its captured knowledge in the form of a directed acyclic causal graph, with nodes being action and state variables and edges denoting the causal relations behind predictions. Furthermore, we design this causal discovery process to be state-dependent, enabling it to model the dynamics in latent causal graphs. Concretely, we conduct causal discovery from the perspective of Granger causality and propose a self-explainable imitation learning framework, {\method}. The proposed framework is composed of three parts: a dynamic causal discovery module, a causality encoding module, and a prediction module, and is trained in an end-to-end manner. After the model is learned, we can obtain causal relations among states and action variables behind its decisions, exposing policies learned by it. Experimental results on both synthetic and real-world datasets demonstrate the effectiveness of the proposed {\method} in learning the dynamic causal graphs for understanding the decision-making of imitation learning meanwhile maintaining high prediction accuracy. 
 
\end{abstract}


%%
%% Keywords. The author(s) should pick words that accurately describe
%% the work being presented. Separate the keywords with commas.

\maketitle

\input{Introduction.tex}

\input{related_work.tex}

\input{Preliminary.tex}

\input{methodology_v1.tex}

\input{experiment.tex}

\input{conclusion.tex}

\begin{acks}
This material is based upon work supported by, or in part by the National Science Foundation (NSF) under grant number IIS-1909702, Army Research Office (ARO) under grant number W911NF-21-1-0198, Department of Homeland Security (DHS) CINA under grant number E205949D, and Cisco Faculty Research Award.
\end{acks}

\bibliographystyle{ACM-Reference-Format}
\balance
\bibliography{file1}

\newpage
\appendix
\input{appendix.tex}

\end{document}

%% file: math_command.tex
\usepackage{amsmath,amsfonts,bm,nccmath}

\def\eqref#1{equation~\ref{#1}}
\def\1{\bm{1}}

\DeclareMathAlphabet{\mathsfit}{\encodingdefault}{\sfdefault}{m}{sl}
\SetMathAlphabet{\mathsfit}{bold}{\encodingdefault}{\sfdefault}{bx}{n}

%% file: Introduction.tex
\section{Introduction}
In imitation learning, neural agents are trained to acquire control policies by mimicking expert demonstrations. It circumvents two vital deficiencies of traditional DRL methods: low sampling efficiency and reward sparsity. Following demonstrations that return near-optimal rewards, the imitator can prevent a vast amount of unreasonable attempts during explorations and has been shown to be promising in many real-world applications~\cite{gu2017deep,jin2018real,zhao2020balancing,ren2022semi,liang2023learn}. However, despite the high performance of imitating neural agents, one problem persists in the interpretability of control policies learned by them. With deep neural networks used as the policy model, the decision mechanism of the trained neural agent is not transparent and remains a black box, making it difficult to trust the model and apply it on high-stake scenarios like the medical domain~\cite{wang2020adversarial,ren2017robust}.

Many efforts have been made to increase the interpretability of policy agents. For example, Reference~\cite{zahavy2016graying} and \cite{mott2019towards} compute saliency maps to highlight critical features using gradient information or attention mechanism; ~\cite{zambaldi2018relational} models interactions among entities via relational reasoning; ~\cite{lyu2019sdrl} designs sub-tasks to make decisions with symbolic planning. However, these methods either provide explanations that are noisy and difficult to interpret~\cite{zahavy2016graying,mott2019towards}, only in the instance level without a global view of the overall policy or make too strong assumptions on the neural agent and lack generality~\cite{lyu2019sdrl}.

%\textit{Can we design a more general framework to provide global interpretations for policy agents?} 
To increase the interpretability of learned neural agents, we propose to explain it from the cause-effect perspective, exposing causal relations among observed state variables and outcome decisions. Inspired by advances in discovering directed acyclic graphs (DAGs)~\cite{zheng2018dags}, we aim to learn a self-explainable imitator by discovering the causal relationship between states and actions. Concretely, taking observable state variables and candidate actions as nodes, the neural agent can generate a DAG to depict the underlying dependency between states and actions, with edges representing causal relationships. For example, in the medical domain, the obtained DAG can contain relations like ``Inactive muscle responses often indicates losing of speaking capability'' or ``Severe liver disease would encourage the agent to recommend using Vancomycin'', as shown in the case study Figure ~\ref{fig:templates}. Such exposed relations can improve user understanding of the policies of the neural agent from a global view, and can provide better explanations of decisions made by it. 

%However, it is a very difficult task due to three main reasons. 1) Decision-making process of neural agents is difficult to interpret. Modern policy models are usually implemented as a deep neural network, in which the utilization of features is entangled and nonlinear, resulting in lack of interpret-ability. 2) The size of candidate causal relations would grow exponentially with the number of variables. Exactly discovering them is NP-hard~\cite{chickering1994learning,chickering1996learning} and intractable in many real-world scenarios. 3) Latent causal structures behind the agent could evolve over time, instead of staying static throughout the trajectory. For example, as indicated in ~\cite{buras2005animal,frausto1998dynamics}, there are multiple stages in sepsis w.r.t disease severity, which would influence efficacy of drug therapies. However, directly incorporating this temporal dynamic element into causal discovery would give too much flexibility in search space, and may lead to over-fitting problem.

However, designing such interpretable imitators from a causal perspective is a very challenging task, mainly due to two reasons: 1) It is non-trivial to identify causal relations behind the decision-making of imitating agents. Modern imitators are usually implemented as a deep neural network, in which the utilization of features is entangled and nonlinear, and lack interpret-ability; and 2) Imitators need to make decisions in a sequential manner, and latent causal structures behind it could evolve over time, instead of staying static throughout the produced trajectory. For example, in a medical scenario, a trained imitator needs to make sequential decisions that specify how the treatments should be adjusted through time according to the dynamic states of the patient. As indicated in ~\cite{buras2005animal,frausto1998dynamics}, there are multiple stages in the states of patients w.r.t disease severity, which would influence the efficacy of drug therapies and result in different treatment policies at each stage. However, directly incorporating this temporal dynamic element into causal discovery would give too much flexibility in search space, and can easily lead to over-fitting.

Targeting at aforementioned challenges, we build our causal discovery objective upon the notion of Granger causality~\cite{granger1969investigating,bressler2011wiener}, which declares a causal relationship $\mathbf{s}_i \rightarrow \mathbf{a}_j$ between variables $\mathbf{s}_i$ and $\mathbf{a}_j$ if $\mathbf{a}_{j}$ can be better predicted with $\mathbf{s}_{i}$ available than not. A causal discovery module is designed to uncover causal relations among variables, and extracted causes are encoded into embeddings of outcome variables before action prediction. %following the notion of Granger causality. The proposed framework is optimized so that state variables predictive towards actions will be identified, providing explanations on decisions of the neural agent. 
Noted that we intervene on the variables used by the agent during prediction and are interested in how its behavior is affected following the notion of Granger causality, and are not discovering causal relations of real-world data.

Concretely, in this work, we propose to design an imitator which is able to produce DAGs providing interpretations on the control policy alongside predicting actions and name it as Causal-Augmented Imitation Learning ({\method}). Identified causal relations are encoded into variable representations as evidence for making decisions. With this manipulation of inputs, we circumvent the onerous analysis of internal structures of neural agents and manage to model causal discovery as an optimization task. Following the observation that the evolvement of causal structures usually follows a stage-wise process~\cite{buras2005animal}, we assume a set of latent templates during designing the causal discovery module which can both model the temporal dynamics across stages and allows for knowledge sharing within the same stage. Consistency between extracted DAGs and captured policies is guaranteed in design, and this framework can be updated in an end-to-end manner. Intuitive constraints are also enforced to regularize the structure of discovered causal graphs, like encouraging sparsity and preventing loops. In summary, our main contributions are:
\begin{itemize}[leftmargin=*]
    \item We study a novel problem of learning dynamic causal graphs to uncover the knowledge captured as well as latent causes behind agent's decisions.
    %We design a framework {\method} to automatically discover the latent dependence structures captured by the policy model in imitation learning. It is an ad-hoc explanation, as obtained DAGs could uncover the knowledge captured as well as latent causes behind agent's decisions. 
    %\item Through modeling the relation of state and action variables as a DAG and incorporating it in the policy model, we manage to discover causal graphs in an end-to-end manner, along with the actor model.
    \item We propose a novel framework {\method}, which is able to learn dynamic DAGs to capture the casual relation between state variables and actions and adopt the DAGs for decision-making in imitation learning;
    %\item To capture both the sharing knowledge of causal relations and its varying elements across time, we propose to model the extraction of DAGs as stage-dependent, with the help of learned templates.
    \item We conduct experiments on synthetic and real-world datasets to demonstrate the effectiveness of {\method} in learning the dynamic DAGs for understanding the decision making of imitation learning meanwhile maintain high prediction accuracy.
\end{itemize}

%% file: related_work.tex
\section{Related Work}

\subsection{Imitation Learning}
Imitation learning is a special case in the reinforcement learning domain, where an agent (policy model) is trained to perform a task and learn a mapping between observations and actions from expert demonstrations~\cite{hussein2017imitation,zhao2023skill}. Such expert knowledge avoids acquiring skills from scratch, and reduces the difficulties of learning in complex and uncertain environments. Imitation learning has been found effective in a wide range of applications, such as human-computer interaction~\cite{zheng2014autonomous,zhao2020balancing,qin2023read,ren2021cross}, self-driving vehicles~\cite{abbeel2007application,codevilla2018end} and robotic arms~\cite{kober2009learning,hsiao2006imitation}.  

Existing imitation learning algorithms can mainly be categorized into two groups, behavior cloning and reinforcement learning with self-defined reward functions~\cite{wang2020adversarial}. Behavioral cloning establishes a direct mapping between states and actions on expert trajectories, hence rewards can be obtained similar to the supervised learning setting~\cite{widrow1964pattern,torabi2018behavioral}. However, it could suffer from reward sparsity problems due to insufficient demonstrations. The other group designs their own reward scores, like inverse reinforcement learning~\cite{ziebart2008maximum} which learns a reward function that would be maximized by expert trajectories, and generative adversarial imitation learning (GAIL)~\cite{ho2016generative} which trains a discriminator to tell those generated by agents apart from expert trajectories. 

Besides these progresses, lack of interpretability is a critical weak point shared by most imitation learning methods. Neural agent is usually implemented as a DNN and treated as a black box. Currently, interpreting imitation learning agents is still an under-explored task. Similar to our idea, Reference~\cite{de2019causal} also introduces causality into imitation learning. However, it identifies causal relations through targeted intervention and feature masking, which is computation extensive and relies upon domain knowledge.

\subsection{Structure Learning from Time-Series}
The task of modeling discrete-time temporal dynamics in DAGs, which is also known as learning dynamic Bayesian networks (DBNs), has generated significant interest in recent years~\cite{khanna2019economy,haufe2010sparse,pamfil2020dynotears,zhao2023towards,liang2023structure,}. It has been used successfully in a variety of domains like disease prognosis~\cite{van2008dynamic} and speech recognition~\cite{meng2018improving,ren2022mitigating}. Most existing approaches are designed as \textit{structural vector autoregressive} (SVAR) models, learning cross-time dependency among variables. For example, Reference~\cite{pamfil2020dynotears} fits a linear structural equation model (SEM) on temporal sequences. Reference~\cite{tank2018neural} designs an LSTM-based model to latently encode SEM, and is able to model nonlinear relations. However, in these works, conditional dependence among variables is taken as stationary across time points. In many real-world applications, the conditional dependency among variables might change. For example, the impacts of oil price on GDP growth are significantly different when the economy is in a high growth phase versus low growth phase~\cite{balcilar2017impact}. Initial efforts have been made to allow the modeling of dynamic DAGs~\cite{tomasi2018latent,hallac2017network} via regularization like graph LASSO. More related to ours is the work~\cite{hsieh2021srvarm}, which learns state-dependent linear SEMs with the assumption of multiple stages.

Our work differentiates from these methods mainly from two perspectives. First, we focus on incorporating causal discovery into the design of imitators to make them more interpretable. Second, we make little assumption on the form of causal models as the decision policy is usually very complex and nonlinear.

%with continuous approximations of the acyclicity constraint~\cite{zheng2018dags,yu2019dag}.
%Methods are designed with assumptions on causal models as additive~\cite{voorman2014graph,buhlmann2014cam}, generalized linear~\cite{zheng2018dags,gu2019penalized}, nonlinear~\cite{tank2018neural,yu2019dag}, etc.

%\subsubsection{Causal Reinforcement Learning}

%% file: Preliminary.tex
\section{Problem Definition}
Throughout this work, we use $\mathcal{S}$ and $\mathcal{A}$ to denote sets of states and actions, respectively. In a classical discrete-time stochastic control process, the state at each time step is dependent upon the state and action from the previous step: $\mathbf{s}_{t+1} \sim \mathbf{P}(\mathbf{s}|\mathbf{s}_{t}, \mathbf{a}_{t})$. $\mathbf{s}_{t} \in \mathcal{S}$ is the state vector in time step $t$, consisting of descriptions over observable state variables. $\mathbf{a}_{t} \in \mathbb{R}^K$ indicates actions taken in time $t$, and $K$ is the size of candidate action set $|\mathcal{A}|$.  Traditionally, deep reinforcement learning dedicates to learning a policy model $\pi_{\theta}$ to select actions given states: $\pi_{\theta}(\mathbf{s}) = \mathbf{P}_{\pi_{\theta}}( \mathbf{a} | \mathbf{s})$, which can maximize long-term rewards. In imitation learning setting, ground-truth rewards on actions at each time step are not available. Instead, a set of demonstration trajectories $\tau$ = $\{\tau_1, \tau_2, ..., \tau_m\}$ sampled from expert policy $\pi_{E}$ is given, where $\tau_i = (\mathbf{s}_0, \mathbf{a}_0, \mathbf{s}_1, \mathbf{a}_1, ... )$ is the $i$-th trajectory with $\mathbf{s}_t$ and $\mathbf{a}_t$ being the state and action at time step $t$. Accordingly, the target is changed to learn a policy $\pi_{\theta}$ that mimics the behavior of expert $\pi_{E}$.  A summary of notations is provided in Appendix.~\ref{ap:notation}.

%Each state-action pair is accompanied by a reward function $\mathcal{R}(\mathbf{s}_{t}, \mathbf{a}_{t})$. The purpose is to learn a decision policy $\pi_{\theta}$ that takes actions in $\mathcal{\mathcal{A}}$ given states in $\mathcal{S}$: $\pi_{\theta}(\mathbf{s}) = \mathbf{P}_{\pi_{\theta}}( \mathbf{a} | \mathbf{s})$.
% that maximize the expected $\gamma$-discounted cumulative reward: 
%\begin{equation}
%    \begin{aligned}
%    \mathbb{E}_{\mathbf{a}_t \sim \pi_{\theta}(\mathbf{s}_t)}&\big{(}\sum_{t=0}^{\infty} \gamma^{t} \mathcal{R}(\mathbf{s}_t, \mathbf{a}_t) \big{)}, \\
%    \text{s.t.} \quad &
%    \mathbf{s}_0 \sim \mathbf{P}_0,  \mathbf{s}_{t+1} \sim \mathbf{P}(\mathbf{s}|\mathbf{s}_{t}, \mathbf{a}_{t})
%    \end{aligned}
%\end{equation}
%Throughout this work, we use $\tau_{\theta}$ and $\rho_{\pi_{\theta}}$ to represent trajectory and state-action pairs generated from policy $\pi_{\theta}$,

In this work, besides obtaining the policy model $\pi_{\theta}$, we further seek to provide interpretations for its decisions. Using notations from the causality scope, we focus on discovering the cause-effect dependency among observed states and predicted actions encoded in $\pi_{\theta}$. Without loss of generality, we can formalize it as a causal discovery task. Concretely, we model causal relations with an augmented linear Structural Equation Model (SEM)~\cite{zheng2018dags}:
\begin{equation}
    \begin{aligned}
    \mathbf{s}_{t+1}, \mathbf{a}_t &= f_2(\mathcal{G}_t \cdot f_1(\mathbf{s}_t,\mathbf{a}_{t-1}))
    \end{aligned}
\end{equation}
In this equation, $f_1, f_2$ are nonlinear transformation functions. Directed acyclic graph (DAG) $\mathcal{G}_{t} \in \mathbb{R}^{(\mathcal{S}+\mathcal{A}) \times (\mathcal{S}+\mathcal{A})}$ can be represented as an adjacency matrix as it is unattributed. $\mathcal{G}_t$ measures the causal relation of state variables $\mathbf{s}$ and action variable $\mathbf{a}$ in time step $t$, and sheds lights on interpreting decision mechanism of $\pi_{\theta}$. It exposes the latent interaction mechanism between state and action variables lying behind $\pi_{\theta}$. The task can be formally defined as: 

\newtheorem{problem}{Problem}
\begin{problem}
\noindent{} \textit{Given $m$ expert trajectories represented as $\tau$, learn a policy model $\pi_{\theta}$ that predicts the action $\mathbf{a}_t$ based on states $\mathbf{s}_t$, along with a DAG $\mathcal{G}_{t}$ exposing the causal structure captured by it in the current time step. This self-explaining strategy helps to improve user understanding of trained imitator.}
\end{problem}

%A complete summary of notations is available in the Appendix ~\ref{append:Notation}.

%% file: methodology_v1.tex
\section{Methodology}
\begin{figure*}[t!]
  \centering
    \includegraphics[width=0.88\textwidth]{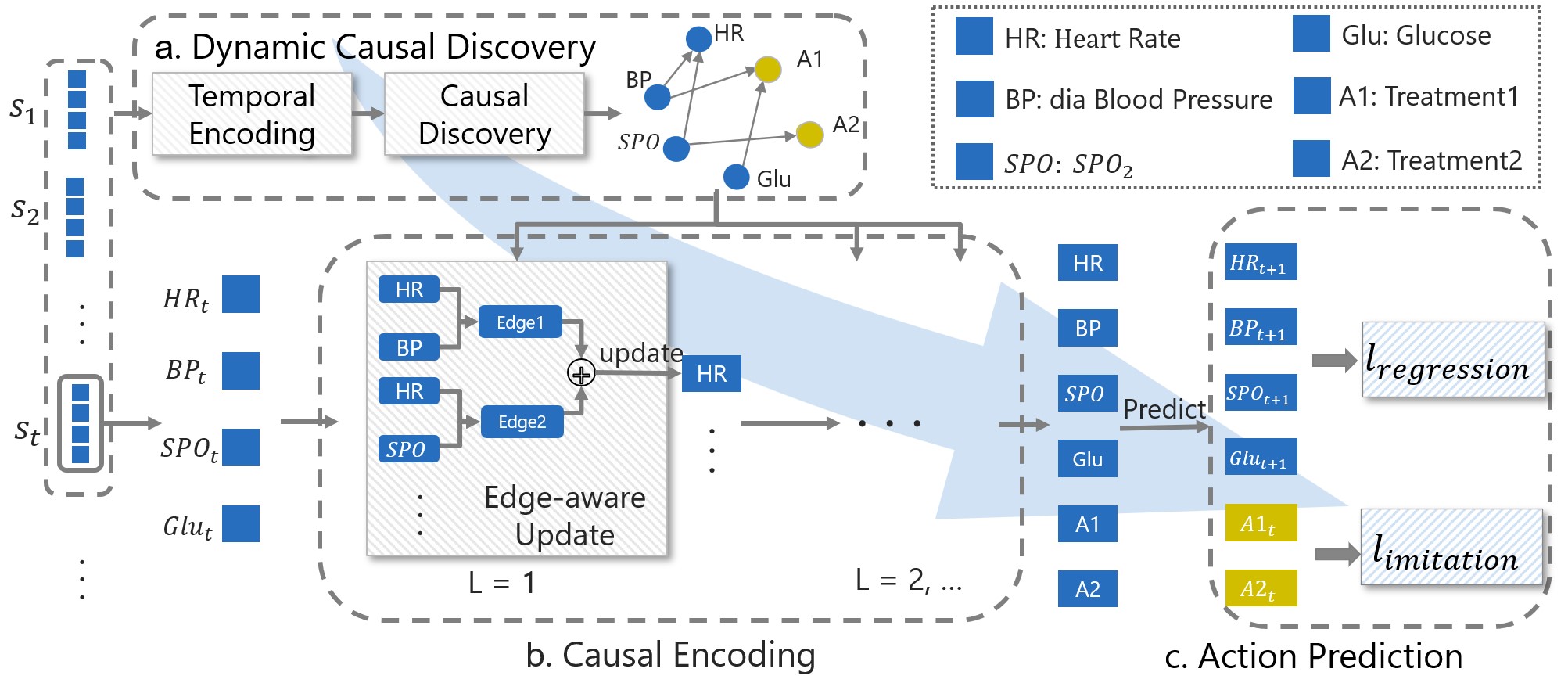}
    \vskip -1.5em
    \caption{Overview of {\method} for interpretable imitation learning from the causality perspective, which is composed of (a) causal discovery module, (b) causal encoding module, (c) prediction module. The arrow in the back shows the inference order of them. } \label{fig:overview}
    %\vskip -1em
\end{figure*}

In this section, we introduce the details of the proposed framework {\method}. The basic idea of {\method} is to discover the causal relationships between state and action and utilize the causal relations to help the agent make decisions. The discovered causal graphs can also provide a high-level interpretation of the neural agent, exposing the reasons behind its decisions. An overview of the proposed {\method} is provided in Figure~\ref{fig:overview}. Concretely, we develop a self-explaining framework that can provide the latent causal graph besides predicted actions, which is composed of: (1) a causal discovery module that constructs a causal graph capturing the causal relations among states and actions for each time step, which can help decision of which action to take next and explain the decision; (2) a causal encoding module which models causal graphs to encode the discovered causal relations for imitation learning; and (3) a prediction module that conducts the imitation learning task based on both the current state and causal relation. All three components are trained end-to-end, and this design guarantees the conformity between discovered causal structures and the behavior of $\pi_{\theta}$. Next, we will introduce the detailed design of these modules one by one.

\subsection{Dynamic Causal Discovery}
%In imitation learning, the policy model $\pi_{\theta}$ is trained to make sequential decisions in a Markov decision process (MDP) utilizing demonstrations from experts. 
%\suhang{@Tianxiang, it's not we use causal graph to represent what policy model discovered. It is we learn causal graph to depict the causal relation and utilize the causal relation for policy making and interpretation.}
Discovering the causal relations between state and action variables can help decision-making of neural agents and increase their interpretability. However, for many real-world applications, the latent generation process of observable states $\mathbf{s}$ and the corresponding action $\mathbf{a}$ may undergo transitions at different periods of the trajectory. For example, there are multiple stages for a patient, such as ``just infected'', ``become severe'' and ``begin to recovery''. Different stages of patients would influence the efficacy of drug therapies~\cite{buras2005animal,frausto1998dynamics}, making it sub-optimal to use one fixed causal graph to model policy $\pi_{\theta}$. On the other hand, separately fitting a ${\mathcal{G}_t}$ at each time step is an onerous task, and could suffer from the lack of training data. %\suhang{1. it is unclear what the causal graph is and why learning causal graph would be helpful. Hope you will make these clear with some examples in the introduction and reemphasize here. 2. Are we going to introduce the concept of stages, i.e., patients might went through different stages, and the causal relationship between symptom and actions (medicines) will remain similar within the same stage. The current description is not clear and can not well explain why we have templates.} 

To address this problem, we design a causal discovery module to produce dynamic causal graphs. Concretely, we assume that the evolving of a time series can be split into multiple stages, and the casual relationship within each stage is static. This assumption widely holds in many real-world applications, as observed in~\cite{buras2005animal,frausto1998dynamics,hallac2017network}. Under this assumption, a discovery model with $M$ DAG templates is designed, and $\mathcal{G}_{t}$ is extracted as a soft selection of those templates.

\subsubsection{Causal Graph Learning}
\begin{figure}[t!]
  \centering
    \includegraphics[width=0.47\textwidth]{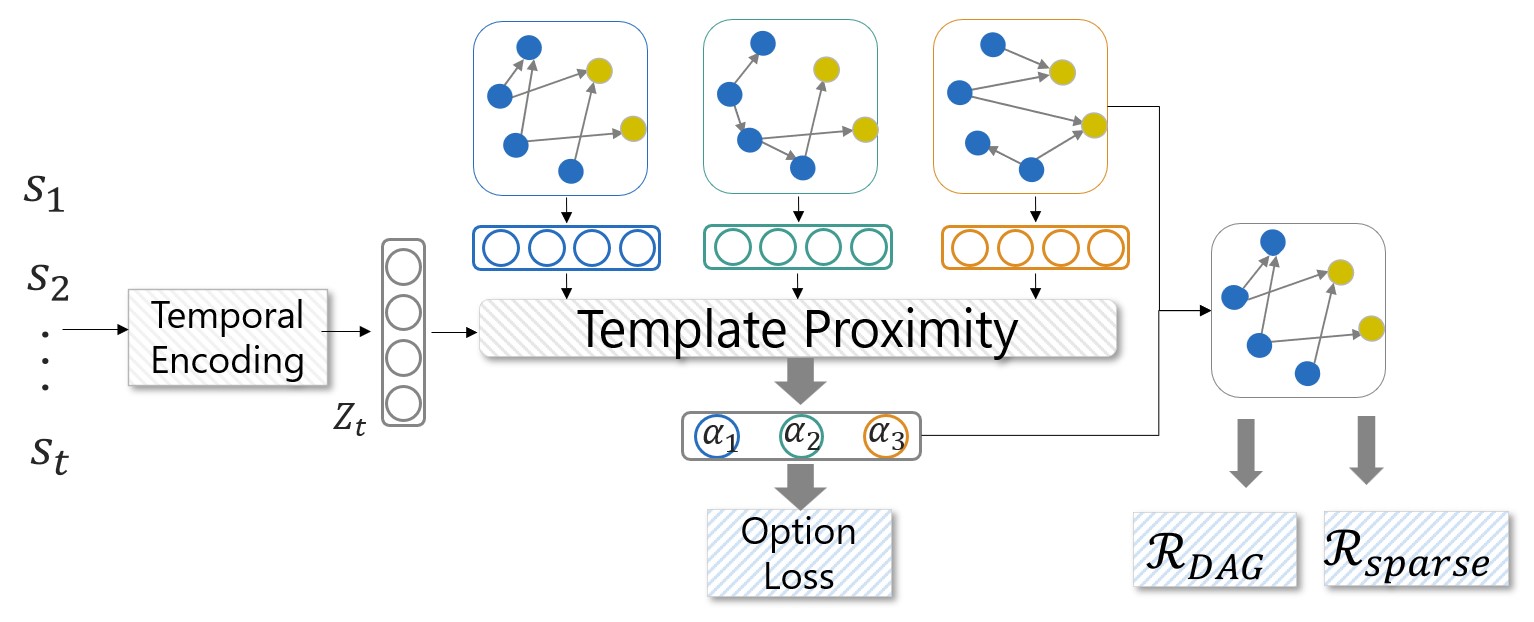}
    \vskip -1.5em
    \caption{Overview of the dynamic causal discovery module.} \label{fig:CausalDiscover}
  \setlength{\abovecaptionskip}{0cm}
      \vskip -1.5em
\end{figure}

An illustration of this causal discovery module is shown in Figure~\ref{fig:CausalDiscover}. Specifically, we first construct an explicit dictionary $\{\mathcal{G}^i, i \in [1,2,...,M]\}$ as the DAG templates. $\mathcal{G}^i \in \mathbb{R}^{(\mathcal{S}+\mathcal{A}) \times (\mathcal{S}+\mathcal{A})}$ and these templates are randomly initialized and will be learned together with the other modules of {\method}. They encode the time-variate part of causal relations. 

Following existing work~\cite{zheng2018dags}, we add the sparsity constraint and the acyclicity regularizer on $\mathcal{G}^i$ to make sure that $\mathcal{G}^i$ is directed acyclic graph. The sparsity regularizer applies $L1$ norm on the causal graph templates to encourage sparsity of discovered causal relations so that those non-causal edges could be removed. It can be mathematically written as
\begin{equation}
    \begin{aligned}
         \min_{\{\mathcal{G}^i,i \in [1,2,...,M]\}} \mathcal{R}_{sparsity} = \sum_{i=1}^{M}|\mathcal{G}^{i}|.
    \end{aligned}
\end{equation}
where $|\mathcal{G}^{i}|$ denotes number of edges inside it. 

In a causal graph, edges are directed and a node cannot be its own descendant. To enforce such constraint on extracted graphs, we adopt the acyclicity regularization in~\cite{yu2019dag}. Concretely, $\mathcal{G}^i$ is acyclic if and only if $\mathcal{H}(\mathcal{G}^i) = tr[e^{\mathcal{G}^i \circ \mathcal{G}^i}] - (|\mathcal{S}| + |\mathcal{A}|) = 0$, where $\circ$ is element-wise square, $e^{\mathbf{A}}$ is the matrix exponential of $\mathbf{A}$, and $tr$ denotes matrix trace. $|\mathcal{S}|$ and $|\mathcal{A}|$ are the number of state and action variables, respectively. Then the regularizer to make the graph acyclic can be written as:
\begin{equation}
    \begin{aligned}\label{eq:acyc}
        \min_{\{\mathcal{G}^i,i \in [1,2,...,M]\}} \mathcal{R}_{DAG} =  \sum_{i=1}^{M}\big(\mathcal{H}(\mathcal{G}^{i}) - (|\mathcal{S}| + |\mathcal{A}|) \big).
    \end{aligned}
\end{equation}
%This regularization term is applied on graph templates. 
When $\mathcal{R}_{DAG}$ is minimized to be $0$, there would be no loops in the discovered causal graphs and they are guaranteed to be DAGs. 

\subsubsection{Causal Graph Selection}
With the DAG templates, at each time stamp $t$, we select one DAG from the templates that can well describe the causal relation between state variables and actions at the current status. Specifically, we use a temporal encoding network to learn the representation of the trajectory for input time step $t$ as 
\begin{equation}
    \begin{aligned}
        \mathbf{z}_t &= Enc(\mathbf{s}_{1}, \mathbf{s}_{2}, ..., \mathbf{s}_t).\\
    \end{aligned}
\end{equation}
In the experiments, we apply a Temporal CNN as the encoding model. Note that other sequence encoding models like LSTM and Transformer can also be used here. For each template $\mathcal{G}^i$, we also learn its representation as:
\begin{equation}
        \mathbf{u}^i = g(\mathcal{G}^{i}).
\end{equation}
As $\mathcal{G}$ is unattributed and its nodes are ordered, we implement $g()$ as an MLP with flattened $\mathcal{G}$ as input, i.e., the connectivity of each node. Note that geometric networks like graph neural networks (GNNs)~\cite{kipf2016semi,Hamilton2017InductiveRL} can also be used here.

Since $\mathbf{z}_t$ captures the trajectory up to time $t$, we can use $\mathbf{z}_t$ to generate $\mathcal{G}_{t}$ by selecting from templates  $\{\mathcal{G}^i\}$ as
\begin{equation} \label{eq:selection_weight}
    \begin{aligned}
        \alpha_{t}^{i} = \frac{\exp(\langle \mathbf{z}_t, \mathbf{u}^i \rangle/T)}{\sum_{i =1}^{M}\exp(\langle \mathbf{z}_t, \mathbf{u}^i \rangle /T)}, \quad \mathcal{G}_{t} = \sum_{i=1}^{M} \alpha_t^i \cdot \mathcal{G}^{i}
    \end{aligned}
\end{equation}
where $\langle, \rangle$ denotes vector inner-product. Here, we adopt a soft selection by setting temperature $T$ to a small value, $0.1$. A small $T$ would make $\alpha_{t}^i$ more close to $0$ or $1$.

%\subsubsection{Regularization}
%Besides gradients from imitation learning task, three regularization terms are designed to guide the extraction of causal graphs. 

To encourage consistency in template selection across similar time steps, we design the template selection regularization loss here. Specifically, states and historical actions at each time are concatenated and clustered into $M$ groups beforehand. We use $q_t^i$ to denote whether time steps $t$ belongs to group $i$, which is obtained from the clustering results. Then, the loss function for guiding the template selection can be written as
\begin{equation}
    \min_{\theta} \mathcal{R}_{option}=-\sum_{i=1}^M \sum_{t} q_t^i log\alpha_t^i.
\end{equation}
where $\alpha_t^{i}$ is the selection weight of time step $t$ on template $i$ from Eq.(\ref{eq:selection_weight}) and  $\theta$ is the set of parameters of graph templates, temporal encoding network $Enc$ and $g()$.

\subsection{Encoding Causal Relations into Embeddings}
For the purpose of learning $\mathcal{G}$ to capture causal structures, we need to guarantee its consistency with the behavior of $\pi_{\theta}$. In this work, we achieve that on the input level. Specifically, we obtain variable embeddings by modeling the interactions among them based on discovered causal relations, and then train $\pi_{\theta}$ on top of these updated embeddings. In this way, the structure of $\mathcal{G}_t$ can be updated along with the optimization of $\pi_{\theta}$. Next, we will introduce the process of encoding causality into variable embeddings in detail.

\subsubsection{Variable Initialization}
Let $\mathbf{s}_{t,j}$ denote state variable $\mathbf{s}_{j}$ at time $t$. First, we map each observed variable $\mathbf{s}_{t,j}$ to the embedding of the same shape for future computations with:
\begin{equation}
    \begin{aligned}
        \hat{\mathbf{h}}_{t,j}^{0} &= \mathbf{s}_{t,j} \cdot \mathbf{E}_{j},
    \end{aligned}
\end{equation}
where $\mathbf{E}_j \in \mathbb{R}^{|\mathbf{s}_j|\times d}$ is the embedding matrix to be learned for the $j$-th observed variable. $\hat{\mathbf{h}}_{t}^{0} \in \mathbb{R}^{|\mathcal{S}| \times d}$, $d$ is the dimension of embedding for each variable. We further extend it to $\mathbf{h}_{t}^{0} \in \mathbb{R}^{(|\mathcal{S}|+|\mathcal{A}|)  \times d}$, to include representation of actions. Representation of these actions are initialized as zero and are learned during training.

\subsubsection{Causal Relation Encoding}
Then, we update the representation of all variables using $\mathcal{G}_{t}$, which aims to encode the casual relation with the representations. In many real-world cases, variables may contain very different semantics and directly fusing them using homophily-based GNNs like GCN~\cite{kipf2016semi} is improper. To better model the heterogeneous property of variables, we adopt an edge-aware architecture:
\begin{equation}
    \begin{aligned}
        \mathbf{m}_{j \rightarrow i} &= [\mathbf{h}_{i,t}^{l-1}, \mathbf{h}_{j,t}^{l-1}] \cdot \mathbf{W}_{edge}^{l} \\
        \mathbf{h}_{i,t}^{l} &= \sigma \big([\sum_{j \in \mathcal{V}}\bar{\mathcal{G}}_{j, i}\mathbf{m}_{j \rightarrow i}, \mathbf{h}_{i,t}^{l-1}]  \mathbf{W}_{agg}^{l} \big)
    \end{aligned}
\end{equation}
where $\mathbf{W}^{l}_{edge}$ and $\mathbf{W}^{l}_{agg}$ are the parameter matrices for edge-wise propagation and node-wise aggregation respectively in layer $l$. $\mathbf{m}_{j \rightarrow i}$ refers to the message from node $j$ to node $i$. In the experiments, $L$ is set as $2$ if not stated otherwise.

\subsection{Prediction with Causality-Encoded Embeddings}
After obtaining causality-encoded variable embeddings, a prediction module is implemented on top of them to conduct the imitation learning task. Its gradients will be back-propagated through the causal encoding module to the causal discovery module, hence informative edges containing causal relations can be identified. In this section, we will introduce the detailed design of this module, along with its training signals.

\subsubsection{Imitation Learning Task}
After previous steps, now $\mathbf{h}_{t,j}$ encodes both observations and causal factors for variable $j$. Then, we make predictions on $\mathbf{a}_t$, which is a vector of length $|\mathcal{A}|$, with each dimension indicating whether to take the corresponding action or not. Concretely, for action candidate $a'$, the process is as follows: (1) $\mathbf{h}_{t,a'}$ and $a'_{t-1}$ are concatenated as the input evidence. $\mathbf{h}_{t,a'}$ is the obtained embedding for variable $a'$ at time $t$, and $a'_{t-1}$ corresponds to the history action from last time. (2) The branch $a'$ of trained policy model $\pi_{\theta}$ predicts the action $a'_{t}$ based on $[\mathbf{h}_{t,a'},a'_{t-1}]$. In our implementation, $\pi_{\theta}$ is composed of $|\mathcal{A}|$ branches with each branch corresponding to one certain action variable. %\suhang{can you write the equation of the prediction?}

Following existing works~\cite{ho2016generative}, the proposed policy model is adversarially trained with a discriminator $D$ to imitate expert decisions. Specifically, the policy $\pi_\theta$ aims to generate realistic trajectories that can mimic $\pi_E$ so as to fool the discriminator $D$; while the discriminator aims to differentiate if a trajectory is from $\pi_\theta$ or $\pi_E$. Through such a min-max game, $\pi_\theta$ can imitate the expert trajectories. The learning objective $\mathcal{L}_{imi}$ on policy $\pi_\theta$ is given as:
\begin{equation}
\begin{aligned}
    \min_{\pi_\theta}& \mathbb{E}_{(\mathbf{s}, \mathbf{a}) \sim \rho_{\pi_\theta}}log(1-D(\mathbf{s},\mathbf{a})) - \lambda H(\pi_{\theta}) \\
    &-\mathbb{E}_{\tau_i \in \tau} \mathbb{E}_{(\mathbf{s}_t, \mathbf{a}_t)\sim \tau_i} P_{\pi_{\theta}}(\mathbf{a}_t \mid \mathbf{s}_t),
\end{aligned}
\end{equation}
where $\rho_{\pi_\theta}$ is the trajectory generated by $\pi_{\theta}$ and $\tau$ is the set of expert demonstrations. $H(\pi) \triangleq \mathbb{E}_{\pi_\theta}[-log\pi(\mathbf{a}|\mathbf{s})]$ is the entropy that encourages $\pi_\theta$ to explore and make diverse decisions. The discriminator $D$ is trained to differentiate expert paths from those generated by $\pi_{\theta}$, whose objective function is:
\begin{equation}
\begin{aligned}
    \max_{D} \mathbb{E}_{\rho_E}log(D(\mathbf{s},\mathbf{a})) + \mathbb{E}_{\rho_{\theta}}log(1-D(\mathbf{s},\mathbf{a})) 
\end{aligned}\label{eq:discriminator}
\end{equation}
Our framework is agnostic towards architecture choices of policy model $\pi_{\theta}$. In the experiments, $\pi_{\theta}$ is implemented as a three-layer MLP, with the first two layers shared by all branches. Relu is selected as the activation function.

\subsubsection{Auxiliary Regression Task}
Besides the common imitation learning task, we further conduct an auto-regression task on state variables. This task can provide auxiliary signals to guide the discovery of causal relations, like the edge from Blood Pressure to Heart Rate in Figure~\ref{fig:overview}. Similar to the imitation learning task, for state variable $s'$ we use $[\mathbf{h}_{t,s'},s'_{t}]$ as the evidence, and use model $\pi_{\phi}$ to predict $s'_{t+1}$ as $\mathcal{L}_{res}$:
\begin{equation}
\begin{aligned}
    \min_{\pi_\phi} -\mathbb{E}_{\tau_i \in \tau} \mathbb{E}_{(\mathbf{s}_t, \mathbf{a}_t)\sim \tau_i} logP_{\pi_{\phi}}(\mathbf{s}_{t+1} \mid \mathbf{h}_{t,\mathbf{s}},\mathbf{s}_{t}),
\end{aligned}
\end{equation}
in which $P_{\pi_{\phi}}$ denotes the predicted distribution of $\mathbf{s}_{t+1}$. 

\subsection{Final Objective Function of \method}
Putting everything together, the final objective function of the proposed {\method} is given as:
\begin{equation}
\begin{aligned}\label{eq:target}
    \min_{\pi_\phi, \pi_\theta} \max_{D}& \mathcal{L}_{imi} 
    + \gamma_1 \cdot \mathcal{L}_{res} + \lambda_1 \cdot \mathcal{R}_{sparse} + \gamma_2 \cdot \mathcal{R}_{option} \\
    & \text{s.t.}  \quad \mathcal{R}_{DAG} = 0.
\end{aligned}
\end{equation}

\iffalse
\begin{equation}
\begin{aligned}
    \min_{\pi_\phi, \pi_\theta} \max_{D}& \mathbb{E}_{(\mathbf{s}, \mathbf{a}) \sim \rho_{\pi_\theta}}log(1-D(\mathbf{s},\mathbf{a})) - \lambda H(\pi_{\theta}) \\
    &-\mathbb{E}_{\tau_i \in \tau} \mathbb{E}_{(\mathbf{s}_t, \mathbf{a}_t)\sim \tau_i} P_{\pi_{\theta}}(\mathbf{a}_t \mid \mathbf{s}_t) \\
    &-\gamma_1 \cdot \mathbb{E}_{(\mathbf{s}_t, \mathbf{a}_t)\sim \tau_i} logP_{\pi_{\phi}}(\mathbf{s}_{t+1} \mid \mathbf{h}_{\mathbf{s},t},\mathbf{s}_{t}) \\
    &+ \lambda_1 \cdot\sum_{i=1}^{M}|\mathcal{G}^{i}| - \gamma_2 \cdot \sum_{i=1}^M \sum_{t} q_t^i log\alpha_t^i. \\
    & \text{s.t.}  \quad \sum_{i=1}^{M}\mathcal{H}(\mathcal{G}^{i}) = 0.
\end{aligned}
\end{equation}
\fi
\noindent where $\lambda_1, \gamma_1$, and $\gamma_2$ are weights of different losses, and the constraint guarantees acyclicity in graph templates. 

To solve this constrained problem in Equation~\ref{eq:target}, we use the augmented Lagrangian algorithm and get its dual form:

\begin{equation}
\begin{aligned}\label{eq:targetAll}
    \min_{\pi_\phi, \pi_\theta} \max_{D}& \mathcal{L}_{imi} 
    + \gamma_1 \cdot \mathcal{L}_{res} + \lambda_1 \cdot \mathcal{R}_{sparse} + \gamma_2 \cdot \mathcal{R}_{option} \\
    & +\lambda_2 \cdot \mathcal{R}_{DAG} + \frac{c}{2} \dot |\mathcal{R}_{DAG}|^2,
\end{aligned}
\end{equation}
where $\lambda_2$ is the Lagrangian multiplier and $c$ is the penalty parameter. The optimization steps are summarized in Algorithm~\ref{alg:Framwork}. Within each epoch, discriminator and the model parameter $\theta, \phi$ are updated iteratively, as shown from line $2$ to line $5$. Between each epoch, we use augmented Lagrangian algorithm to update the multiplier $\lambda_2$ and penalty weight $c$ from line $6$ to line $11$. These steps progressively increase the weight of $\mathcal{R}_{DAG}$, so that it will gradually converge to zero and templates will satisfy the DAG constraint.

\begin{algorithm}[t]
  \caption{Full Training Algorithm}
  \label{alg:Framwork}
  %\scalebox{0.75}{
  \begin{algorithmic}[1] 
  \REQUIRE %??????????Input
    Demonstrations $\tau$ generated from expert policy $\pi_{E}$, initial template set $\{\mathcal{G}^i, i \in [1,2,...,M]\}$, initial model parameter $\theta, \phi$, hyperparameters $\lambda_1, \lambda_2, \gamma_1, \gamma_2, c$, initialize $\mathcal{H}_{old}=\inf$, parameter in Augmented Lagrangian: $\sigma=\frac{1}{4}$, $\rho=10$
    \WHILE {Not Converged}
    \FOR{$\tau_i \sim \tau$}
    \STATE Update parameter of discriminator $D$ to increase the loss of Equation~\ref{eq:discriminator};
    \STATE Update $\theta,\phi$ with gradients to minimize Equation~\ref{eq:target};
    \ENDFOR
    \STATE Compute $\mathcal{H}$ with Equation~\ref{eq:acyc};
    \STATE $\lambda_2 \leftarrow \lambda_2 + \mathcal{H}\cdot c$
    \IF {$\mathcal{H} \leq \sigma \cdot \mathcal{H}_{old}$}
        \STATE $c \leftarrow c*\rho$
    \ENDIF
    \STATE $\mathcal{H}_{old} \leftarrow \mathcal{H}$
    \ENDWHILE
    
  \RETURN Learned templates $\{\mathcal{G}^i, i \in [1,2,...,M]\}$, trained policy model $\pi_\theta$
  \end{algorithmic}
  %}%% resizebox
\end{algorithm}

%% file: experiment.tex
\section{Experiment}
In this section, we evaluate the prediction accuracy and interpretability of the proposed {\method} on both synthetic and real-world datasets. Specifically, we aim to answer the following questions:
\begin{itemize}[leftmargin=*]
    \item \textbf{RQ1}: Can the proposed approach correctly identify causal relations among state and action variables?
    \item \textbf{RQ2}: Can our proposed method achieve better interpretability without sacrificing performance in imitation learning?
    \item \textbf{RQ3}: How would different hyperparameter configurations influence the effectiveness of proposed method?
\end{itemize}

\subsection{Baselines}
To the best of our knowledge, there is no existing work on discovering DAGs to help learn and interpret imitation learning models. To evaluate capacity of {\method} in identifying causal relations, we compare it with representative and state-of-the-art causal discovery methods in time-series data: (1) cMLP/cLSTM~\cite{tank2018neural}, which discovers nonlinear causal relations by training an MLP or an LSTM for each effect, and its causal factors are identified through analyzing non-zero entries in the weight matrix. (2) SRU/eSRU~\cite{khanna2019economy}, which extends cLSTM through training a component-wise time-series predictor based on Statistical Recurrent Units(SRU). (3) DYNOTEARS~\cite{pamfil2020dynotears}, which designs a score-based approach to capture causal structures in time series. Causal graph is explicitly parameterized. and causal relations are assumed to be linear. (4) SrVARM~\cite{hsieh2021srvarm}, which is similar to DYNOTEARS in design, but assumes that causal graphs are state-specific and fall into $K$ groups. (5) ACD~\cite{lowe2020amortized}, which trains a single amortized model that can infer causal relations across instances with different underlying causal graphs. 

The comparison between our approach and these baselines is summarized in Table~\ref{tab:baseline} in terms of three dimensions: whether discovers dynamic causal relations, whether supports nonlinear causal relations, and whether guarantees acyclicity in discovered causal graphs. Following their designs, we train them through conducting auto-regression directly on expert trajectories $\pi_{E}$.

\subsection{Datasets}
To evaluate the performance of causal discovery, we conduct experiments on a publicly available synthetic dataset Kuramoto~\cite{lowe2020amortized} and a real-world dataset MIMIC-IV~\cite{wang2020adversarial}. To examine the performance of {\method} in imitation learning task ad answer RQ2, we further test it on three classical gym datasets~\cite{gym_minigrid}: MiniGrid-FourRoom, LavaGap, and DoorKey. Descriptions of these datasets are provided in Ap.~\ref{ap:dataset}.

\begin{table}[t!]
  \setlength{\tabcolsep}{4.5pt}
  \normalsize
  \caption{Comparison with Baseline Causal Discovery Models} \label{tab:baseline}
  \vskip -1em
  \begin{tabular}{c   c c c}
    \hline
    Models & Dynamic & Nolinear Causality & Acyclicity \\
    \hline
    cMLP & $\times$ & $\surd$ & $\times$ \\
    cLSTM & $\times$ & $\surd$ & $\times$ \\
    SRU & $\times$ & $\surd$ & $\times$\\
    eSRU &  $\times$ & $\surd$ & $\times$\\
    DYNOTEARS &  $\times$ & $\times$ & $\surd$ \\
    SrVARM &  $\surd$ & $\times$ & $\surd$ \\
    ACD & $\surd$ & $\surd$ & $\times$\\
    \hline
    Our Approach & $\surd$ & $\surd$ & $\surd$ \\
    \hline
  \end{tabular}
  \vskip -1em
\end{table}

\subsection{Configurations}
\subsubsection{Hyperparameter Settings}
For all approaches, the learning rate is initialized as $0.005$ and maximum epoch is set as $1,000$. For both our approach and baselines, we use grid search to find hyper-parameters with the best performance on each dataset. For synthetic dataset $Kuramoto$, $M$ is fixed as $3$ for ease of evaluation. For real-world datasets, Sepsis and Comorbidity, $M$ is also set as $3$ with prior knowledge on degree of severity~\cite{buras2005animal}. Train:validation:test ratio is split as $2:3:5$.

\subsubsection{Evaluation Metrics}
To evaluate the quality of discovered causal relations provided by the policy model, we use AUC-ROC score to measure their alignment with the ground-truth causal relationships. A higher AUC-ROC score means more accurate causal discovery performance, which indicates better interpretability. Besides, to evaluate performance in decision making, we also conduct teacher-forcing test and report the accuracy in action prediction.

\begin{table}[t!]
  \setlength{\tabcolsep}{4.5pt}
  \normalsize
  \caption{Causal discovery performance on static Kuramoto dataset with different number of oscillators.} \label{tab:static}
  \vskip -1em
  \begin{tabular}{c  c c c}
    \hline
     & \multicolumn{3}{c}{AUROC}  \\
    Models & Kura5 & Kura10 & Kura50 \\
    \hline
    cMLP & $0.52\pm0.01$ & $0.48\pm0.01$ & $0.49\pm0.02$ \\
    cLSTM & $0.47\pm0.01$ & $0.49\pm0.01$ & $0.49\pm0.01$ \\
    SRU & $0.81\pm0.04$ & $0.64\pm0.02$ & $0.53\pm0.03$ \\
    eSRU & $0.61\pm0.04$ & $0.54\pm0.03$ & $0.51\pm0.02$ \\
    DYNOTEARS & $0.63\pm0.01$ & $0.57\pm0.01$ & $0.55\pm0.02$ \\
    SrVARM & $0.67\pm0.02$ & $0.66\pm0.04$ & $0.57\pm0.03$ \\
    ACD &  $0.82\pm0.04$ & $0.71\pm0.03$ & $0.63\pm0.02$ \\
    \hline
    Ours & $\mathbf{0.95}\pm0.02$ & $\mathbf{0.98}\pm0.01$ & $\mathbf{0.91}\pm0.03$ \\
    \hline
  \end{tabular}
  %\vskip -1em
\end{table}

\subsection{Performance on Discovering Static DAG}
First, we compare the performance of DAG learning of {\method} with baselines in the static setting. Each method is trained until converges to make a fair comparison. We conduct each experiment for $5$ times. Both the average performance and the standard deviation are reported in Table~\ref{tab:static}. From the result, we can observe that:
\begin{itemize}[leftmargin=*]
    \item Our proposed framework achieves the best performance compared to baselines, which validates its capacity to correctly identify nonlinear causal relations;
    \item Our proposed framework scales well on this Kuramoto dataset. Compared to baselines like SRU, its performance is much more stable w.r.t graph sizes.
    \item Due to limitations like only support linear causal relations~\cite{hsieh2021srvarm,pamfil2020dynotears} or unable to guarantee acyclicity~\cite{lowe2020amortized,tank2018neural}, baselines fail to achieve satisfactory performance on this task.
\end{itemize}

\begin{table}[t!]
  \setlength{\tabcolsep}{4.5pt}
  \normalsize
  \caption{Causal discovery performance on dynamic Kuramoto dataset with a different number of oscillators. } \label{tab:dynamic}
  \vskip -1em
  \begin{tabular}{c  c c c}
    \hline
     & \multicolumn{3}{c}{AUROC}  \\
    Models & Kura5\_vary & Kura10\_vary & Kura50\_vary \\
    \hline
    SRU-D & $0.78\pm0.03$ & $0.62\pm0.02$ & $0.54\pm0.01$ \\
    DYNOTEAR-D & $0.62\pm0.03$  & $0.57\pm0.02$ & $0.52\pm0.03$ \\
    \hline
    SrVARM & $0.63\pm0.04$  & $0.64\pm0.05$ & $0.56\pm0.02$\\
    ACD & $0.73\pm0.03$  & $0.68\pm0.02$ & $0.58\pm0.04$ \\
    \hline
    Ours & $\mathbf{0.82}\pm0.05$ & $\mathbf{0.78}\pm0.04$ & $\mathbf{0.72}\pm0.03$ \\
    \hline
  \end{tabular}
  \vskip -1em
\end{table}

\subsection{Performance on Discovering Dynamic DAG}
Based on the results on static DAG, we select SRU, DYNOTEARS, SrVARM, and ACD as baselines for the scenario with varying DAGs. For SRU and DYNOTEARS, they are designed only for static causal relation discovery. Hence, we take ground-truth DAG index as known for these baselines and train them once for each DAG. We mark them as SRU-D and DYNOTEAR-D respectively. Other approaches can cope with dynamic causal relations and do not require such modification. The results are summarized in Table~\ref{tab:dynamic}. From the results, we can observe that: (\textbf{i}) Our approach again achieves the best performance, outperforming all baselines with a clear margin; and (\textbf{ii}) It is more difficult to conduct causal discovery when the latent DAGs are dynamic. The performance of all approaches degrades in this setting.
% \begin{itemize}
%     \item Our approach again achieves the best performance, outperforming all baselines with a clear margin.
%     \item It is more difficult to conduct causal discovery when the latent DAGs are dynamic. The performance of all approaches degrades in this setting.
% \end{itemize}

\subsection{Performance on Decision Making}
To answer RQ2, we compare proposed approach with representative methods in imitation learning, including Behavior Cloning (BC), Adversarial Inverse Reinforcement Learning (AIRL)~\cite{fu2018learning} and GAIL~\cite{ho2016generative}. Performance of our backbone model architecture is also reported as Vanilla. For the vanilla model, there is no causal discovery or causal encoding modules, and the policy model predicts actions based on the full observed states. Train, validation and test sets are split as $3:2:5$. Trained policy models are tested on expert trajectories. For Kuramoto dataset we report MSE difference in predicted trajectories (smaller is better), and for Mimic-IV datasets we report the mean Accuracy and AUROC score (higher is better). For three gym datasets, we report the mean accuracy, mean reward, and macro-F score (higher is better). Results are summarized in Table~\ref{tab:imitation_kuramoto}, ~\ref{tab:imitation_mimic4}, and Table~\ref{tab:imitation_gym}. From the three tables, we can observe that the proposed approach can provide an explanation with comparable performance in decision making.

% \begin{table}[h!] 
%   \setlength{\tabcolsep}{4.5pt}
%   \normalsize
%   \caption{Comparison between vanilla imitation learning (IL) and our proposed model on action predictions}
%   %\vskip -1em
%   \subtable[Comparison on Kuramoto dataset, measured using prediction distance. Smaller is better]{
%   \begin{tabular}{c  c c  c c  }
%     \hline
%     Methods & Kura5 & Kura10 & Kura5\_vary & Kura10\_vary \\
%     \hline
%     %Sepsis &  &  &  & \\
%     Imitation Learning & $0.0053$ & $0.0065$ & $0.0122$ & $0.0114$  \\
%     Ours & $0.0056$  & $0.0063$ & $0.0125$  & $0.0115$ \\
%     \hline
%   \end{tabular}
%   }
% \subtable[Comparison on Comorbidity and Sepsis settings of Mimic4 dataset, measured using action prediction performance. Higher is better]{
%   \begin{tabular}{c  c c | c c  }
%     \hline
%     Datasets & IL & Ours & IL & Ours \\
%      \hline
%      & \multicolumn{2}{c}{Micro AUC}  & \multicolumn{2}{|c}{Accuracy}\\
%      \hline
%     Comorbidity & $0.8793$  & $0.8774$  & $0.9618$  & $0.9628$  \\
%     Sepsis & $0.9402$ & $0.9317$ & $0.9818$ & $0.9817$ \\
%     \hline
%   \end{tabular}
%   }
% \label{tab:imitation}
% \end{table}

\begin{table}[t!] 
  %\small
  \normalsize
  \caption{Action prediction performance measured in terms of  prediction distance on Kuramoto datasets. Smaller is better}\label{tab:imitation_kuramoto}
  \vskip -1em
  \begin{tabular}{c  c c  c c  }
    \hline
    Methods & Kura5 & Kura10 & Kura5\_vary & Kura10\_vary \\
    \hline
    %Sepsis &  &  &  & \\
    BC &  $0.0061$ & $0.0069$ & $0.0129$ & $0.0117$ \\
    AIRL & $0.0062$ & $0.0068$ & $0.0133$ & $0.0119$ \\
    GAIL & $0.0059$ & $0.0067$ & $0.0131$ & $0.0128$ \\
    \hline
    Vanilla & $0.0053$ & $0.0065$ & $0.0122$ & $0.0114$  \\
    Ours & $0.0056$  & $0.0063$ & $0.0125$  & $0.0115$ \\
    \hline
  \end{tabular}
  
  \vskip 0.5em
  %\small
  \normalsize
  \centering
  \caption{Action prediction performance measured in terms of  AUC and Accuracy on MIMIC-IV datasets. Higher is better}\label{tab:imitation_mimic4}
  \vskip -1em
  \begin{tabular}{c  c c | c c  }
    \hline
    & \multicolumn{2}{c}{ Comorbidity}  & \multicolumn{2}{|c}{Sepsis}\\
     \hline
    Methods & AUC  & ACC & AUC  & ACC \\
     \hline
    BC & $0.8695$ & $0.9507$  & $0.9311$ & $0.9735$ \\
    AIRL & $0.8574$ & $0.9427$ & $0.9216$ & $0.9673$ \\
    GAIL & $0.8716$ & $0.9585$ & $0.9315$ & $0.9774$ \\
    \hline
    Vanilla & $0.8793$  & $0.9618$  & $0.9402$  & $0.9818$  \\
    Ours & $0.8774$ & $0.9628$ & $0.9317$ & $0.9817$ \\
    \hline
  \end{tabular}
  \vskip -1em
\end{table}

\begin{table*}[h]
    \centering
    \normalsize
  \caption{Action prediction performance measured on three gym datasets. Besides accuracy and MacroF, we also test the trained agent on the environment to obtain the mean rewards.
  }\label{tab:imitation_gym}
  \vskip -1em
    \begin{tabular}{c|c c c | c c c | c c c}
    \hline
    & \multicolumn{3}{c}{ FourRoom}  & \multicolumn{3}{|c}{LavaGap} & \multicolumn{3}{|c}{DoorKey}\\
     \hline
    Methods & ACC & Reward  & macroF  & ACC & Reward  & macroF  & ACC & Reward  & macroF \\
     \hline
     BC & $0.9012$ & $0.2836$ & $0.6737$ & $0.8745$ & $0.9357$ & $0.8634$ & $0.8859$ & $0.8403$ & $0.8618$ \\
     AIRL & $0.8725$ & $0.2617$ & $0.6593$ & $0.8733$ & $0.9342$ & $0.8629$ & $0.8915$ & $0.8672$ & $0.8728$ \\
     GAIL & $0.8468$ & $0.1904$ & $0.6336$ & $0.8659$ & $0.9223$ & $0.8605$ & $0.8733$ & $0.8328$ & $0.8531$  \\
     \hline
     Vanilla & $0.9037$ & $0.2872$ & $0.6795$ & $0.8793$ & $0.9464$ & $0.8728$ & $0.9035$ & $0.8742$ & $0.8813$ \\
     Ours & $0.9045$ & $0.2863$ & $0.6746$ & $0.8821$ & $0.9487$ & $0.8745$ & $0.9012$ & $0.8756$ & $0.8804$ \\
     \hline
    \end{tabular}
    \vskip -1.5em
\end{table*}

\subsection{Ablation Study}
\subsubsection{Analyzing Sparsity Regularization}
\begin{figure}[t!]
%\vskip -1em
  \subfigure[Kura10,Kura10\_vary]{
		\includegraphics[width=0.21\textwidth]{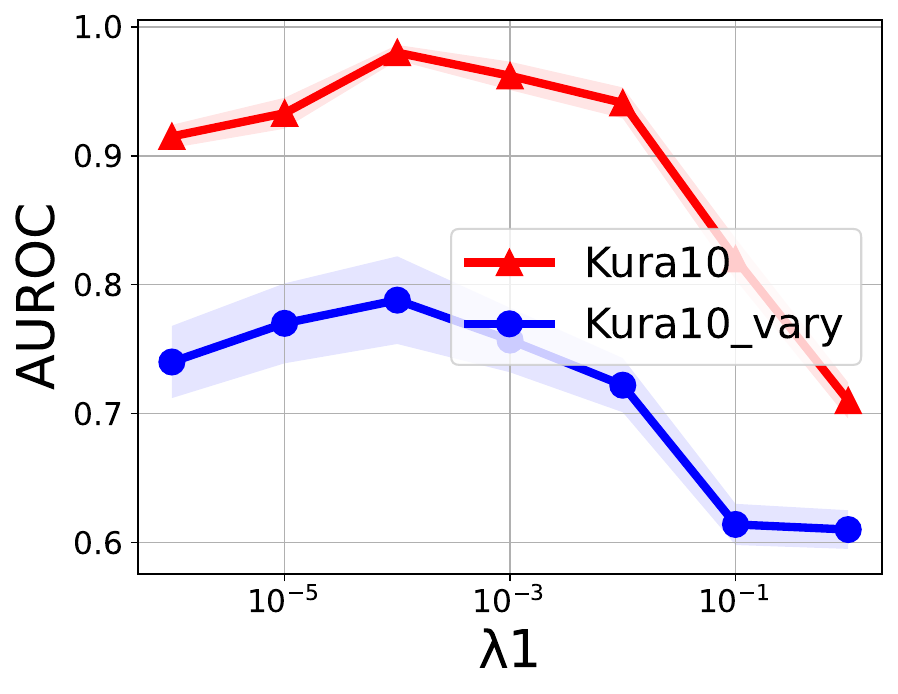}}
  \subfigure[LavaGap, FourRoom]{
		\includegraphics[width=0.21\textwidth]{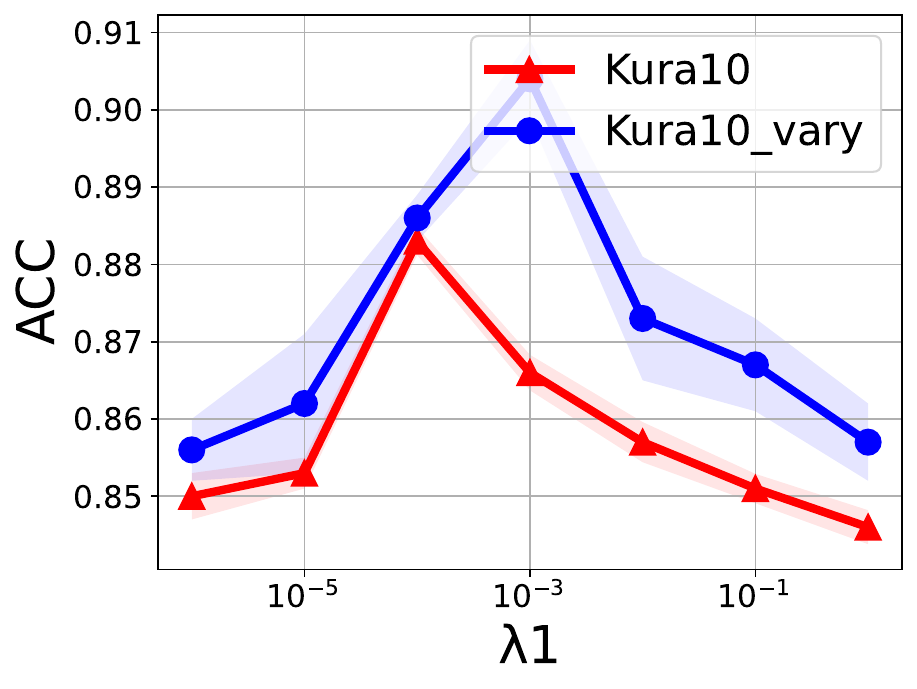}}
		
    \vskip -1.5em
    \caption{Sensitivity on weight of sparsity regularization, $\lambda_1$}
		\label{fig:spar}
    \vskip -1.5em
\end{figure}

In this subsection, we analyze the sensitivity of the proposed {\method} on hyperparameters $\lambda_1$. $\lambda_1$ controls the importance of sparsity regularization term.  We vary it as $\{ 10^{-6}, 10^{-5},\dots, 0.1, 1\}$. Evaluations are conducted both on causal discovery (Kura10 and Kura10\_vary) and on action prediction (LavaGap and FourRoom), with other configurations remaining the same as main experiment. Each experiment is conducted $3$ times, and the average results in causal discovery are shown in Fig~\ref{fig:spar}. From the figure, we can observe that the proposed {\method} performs relatively stable with $\lambda_1 \in [10^{-5}, 10^{-3}]$. Setting $\lambda_1$ to a too-large value, e.g, larger than $0.1$, will result in a sharp drop in the quality of identified causal edges and prediction accuracy. This is because sparsity constraint can encourage the causal discovery module to remove uninformative edges, but setting it too large would remove correctly identified edges as well.

\subsubsection{Analyzing Acyclicity Regularization}
In this subsection, we analyze the sensitivity of {\method} on the importance of acyclicity regularization. Results are presented in Ap.~\ref{ap:sensitivity}.

\subsubsection{Influence of Training Instance Amount}
\begin{figure}[t]
  \centering
  \subfigure[Kura10,Kura10\_vary]{
		\includegraphics[width=0.23\textwidth]{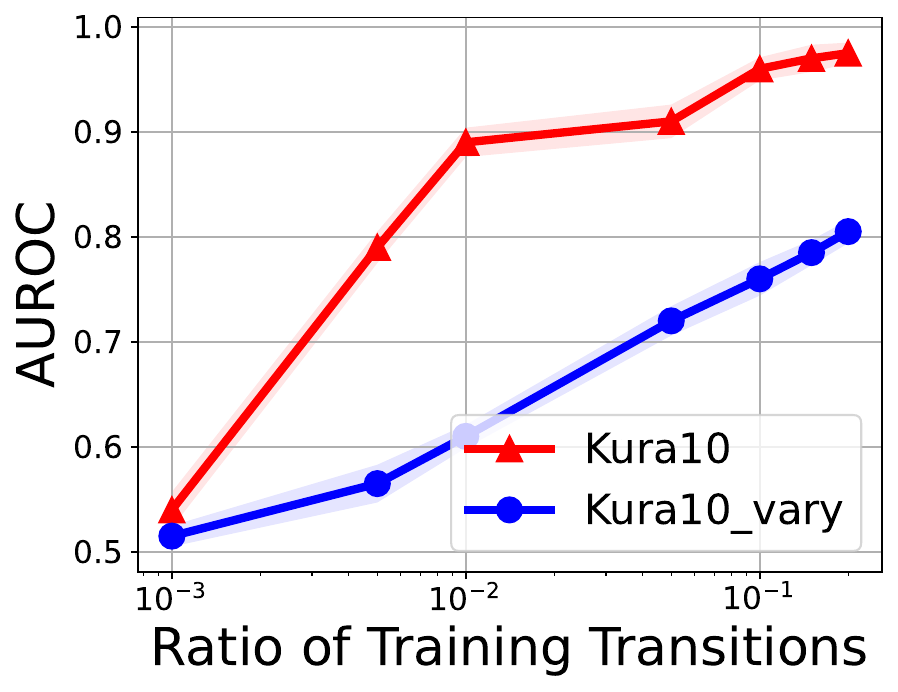}}
  \subfigure[LavaGap, FourRoom]{
		\includegraphics[width=0.23\textwidth]{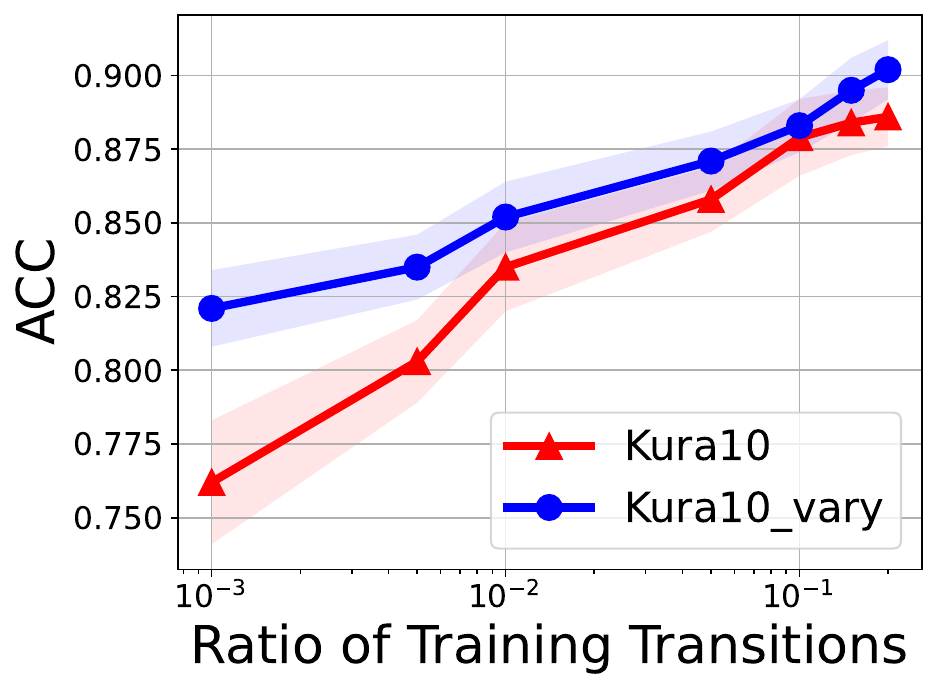}}
		
    \vskip -1em
    \caption{Influence of training instance amount. } \label{fig:src_num}
    \vskip -1em
\end{figure}

In this subsection, we evaluate the influence of training size on causal discovery and action prediction quality, to obtain an idea of the amount of data needed for a successful learning process. We vary the ratio of instances for training as $\{1e-3, 5e-3, 1e-2, 5e-2, 0.1, 0.15, 0.2\}$, and experiment on datasets Kura10, Kura10\_vary, LavaGap and FourRoom. The results are summarized in Fig~\ref{fig:src_num}. From the figure, we can observe that for the Kuramoto dataset, the benefit of increasing training examples is more clear with an amount less than $0.01$ (corresponds to $1,000$ transitions) in the static setting. In the dynamic setting, on the other hand, more training examples are needed for successful training. For datasets LavaGap and FourRoom, the performance become more stable with ratio larger then $0.1$.

\begin{table}[t]
  \setlength{\tabcolsep}{4.5pt}
  \normalsize
  \caption{Causal discovery performance in terms of AUCROC with different number or type of GNN layers used in enforcing causal graphs to the decision-making process.} \label{tab:layers}
  \vskip -1em
  \begin{tabular}{c |  c c | c c}
    \hline 
     %& \multicolumn{4}{c}{AUROC Score}  \\
     & \multicolumn{2}{c|}{Kura10} &\multicolumn{2}{c}{Kura10\_vary} \\
    Layer Number & GCN & Edge-aware & GCN & Edge-aware \\
    \hline
    1-layer & $0.95$ &$0.96$ & $0.77$ & $0.76$ \\
    2-layer &$0.71$ & $0.98$ & $0.64$ & $0.78$ \\
    3-layer &$0.58$ & $0.59$ & $0.56$ & $0.59$ \\
    \hline
  \end{tabular}
  \vskip -1em
\end{table}

\subsubsection{Influence of Model Architectures}
Causal encoding module uses discovered causal graphs to guide the message propagation among variables, and back-propagates gradients to the causal discovery module so that the framework can be trained end-to-end jointly. In this subsection, we analyze the model's performance w.r.t different architectures of it. Particularly, we test two GNN layers: GCN~\cite{kipf2016semi} which is one of the most popular GNN layers, and Edge-aware layer which used in the main experiments. Different numbers of GNN layers are also tested here. Experiments are conducted on Kura10 and Kura10\_vary, with AUROC score on discovered causal edges reported. Results in causal discovery are summarized in Table~\ref{tab:layers}. From the table, we observe that the performance drops quickly as the GCN  goes deep, while the edge-aware layer used in this work performs well for both 1-layer and 2-layer settings. We attribute this observation to the ``homophily'' assumption of GCN, which is ineffective in modeling complex interactions of causal graphs.

\subsection{Case Study}\label{sec:case}

We further conduct case studies on dataset Kuramoto and Mimic-IV. Due to space limitations, we put it in Appendex.~\ref{ap:case1} and ~\ref{ap:case}.

%% file: conclusion.tex
\section{Conclusion}
In this work, we integrate causal discovery into imitation learning and propose a framework with improved interpretability. Besides learning control policies, the trained imitator is able to provide DAGs depicting captured dependence among state and action variables. With dynamic causal discovery module and the causality encoding module implemented as GNNs, the framework can model complex nonlinear causal relations. Experimental results on both simulation and real-world datasets show the effectiveness of the proposed method in capturing the casual relations for explanation and prediction.  There are several interesting directions need further investigation. First, in this paper, we use clustering algorithm to cluster the states into stages and utilize it to supervise template selection. We would like to extend {\method} to learn the stages instead of relying on pre-clustered stages. Second, the identified causal relations could expose distribution shifts across domains. It is promising to utilize them for a more efficient transfer learning algorithm.

%% file: appendix.tex
\section{Notations}\label{ap:notation}
The definition and form of notations used in this work are summarized in Table~\ref{tab:notation}.

\begin{table*}[t]
    \centering
    \caption{Notations}
    \label{tab:notation}
    \vskip -1em
    \begin{tabular}{c|p{13cm}}
    \hline 
         Notation & Description \\ \hline
        $\mathbf{s}_t \in \mathcal{S}$ & State vector in time step $t$. It consists of descriptions of observable variables in $\mathcal{S}$.\\
        $\mathbf{a}_t \in \mathcal{A}$ & Action adopted in time step $t$. $\mathcal{A} \in \{0, 1\}^K$, with $k$-th dimension indicating whether $k$-th action is taken. \\
        $\pi_E$ & The policy taken by experts. It is used as the target policy which the policy model learns to imitate. \\
        $\pi_{\theta}(\mathbf{s})$ & The policy learned by the policy model, which is parameterized using $\theta$. It predicts the action to be taken based on the current state, $\pi_{\theta}(\mathbf{s}) = \mathbf{P}_{\pi_{\theta}}( \mathbf{a} | \mathbf{s})$ \\
        $\tau_i = (\mathbf{s}_0, \mathbf{a}_0,  ... )$ & The $i$-th trajectory generated from expert policy $\pi_{E}$. In total, we assume that $m$ trajectories are available for the learning process. \\
        $\tau_{\theta} = (\mathbf{s}_0, \mathbf{a}_0,  ... )$ &  Trajectory generated from policy model $\pi_{\theta}$. \\
        $\rho_{\pi}$ & The distribution of state-action pairs generated from interaction between policy $\pi$ and the environment.\\% calculated as $\rho_{\pi} = \pi(\mathbf{a}|\mathbf{s})\sum_{t=0}^{\infty}{\color{red}\gamma}P(\mathbf{s}_t=\mathbf{s} \mid \pi)$. \\
        %${\color{red}\mathbf{W}_{p,t}} \in \{0,1\}^{\mathcal{S}}$ & Indicates the causal elements of observed variables in $\mathcal{S}$ as evidence for $\pi_{\theta}$ to make decisions in time step $t$. During training, we make it continuous, $\mathbb{R}^{\mathcal{S}}$, for the ease of optimization. \\
        $\mathcal{G}_{t} \in \mathbb{R}^{(\mathcal{S}+\mathcal{A}) \times (\mathcal{S}+\mathcal{A})}$ &  A DAG representing the latent causal graph among state and action variables in time step $t$. \\ \hline 
        %$\mathbf{W}_{r,t} \in \mathbb{R}^{(\mathcal{S}+\mathcal{A}) \times (\mathcal{S})} $ & Indicates the causality relation from state/action pairs in time step $t$ to time step $t+1$. \\
    \end{tabular}
\end{table*}

\section{Datasets}\label{ap:dataset}

\subsection{Kuramoto}
This dataset contains $1$-D time-series of phase-coupled oscillators, and is used to describe synchronization. Through manipulating the coupling factors among them, we are able to have control upon ground-truth causal graphs. Causal effects over oscillator motions are non-linear in this dataset, which increases the difficulty in discovering them. This simulation dataset enables us to conduct experiments on multiple different settings, and evaluate discovered causal graphs by comparing them with ground-truths.

By default, causal graph remains static across the data generation process. Alternatively, we consider a dynamic scenario by making it switch inside a pre-defined candidate set. The size of candidate set is set to $3$. %A detailed description on dataset construction is provided in Appendix~\ref{sec:Kuramoto}. 
Specifically, we create datasets of three sizes to evaluate the scalability of proposed approach:
\begin{itemize}[leftmargin=0.1in]
    \item Kura5/Kura5\_vary. It contains $5$ oscillators. Four oscillators are used as state variables, and one as the action. Kura5 denotes the static setting, in which the causal graph is fixed, while Kura5\_varying represents the dynamic setting.
    \item Kura10/Kura10\_vary. It contains $10$ oscillators. Eight oscillators are used as state variables, and two as actions.
    \item Kura50/Kura50\_vary. It contains $50$ oscillators. Forty-two oscillators are used as state variables, and eight as actions.
\end{itemize}
Each dataset contains $500$ sequences with random initialization, and each sequence has $100$ time steps. %A more detailed description is provided in the Appendix.

\subsection{MIMIC-IV}
We also conduct experiments on a real-world medical dataset MIMIC-IV, which contains medical records for over $40,000$ patients admitted to intensive care units at BIDMC~\cite{johnson2020mimic}. We gleaned patients diagnosed with Sepsis ($6,620$ patients) and Comorbidity ($8,284$ patients) as two datasets, and select different antibacterial drugs as actions. Following the pre-processing in ~\cite{wang2020adversarial}, each sepsis record contains $8$ symptoms and comorbidity record contains $11$ symptoms, and $14$ treatments/drugs are used as actions for both datasets. The neural agent is trained to recommend treatments based on diagnosed symptoms.  In this dataset, there is no ground-truth causal relations available, making it difficult to evaluate knowledge learned by the imitator. Addressing this, we have human experts (i.e., doctors) help us define some rules according to pathogenesis of sepsis~\cite{li2020temporal} and design a set of case studies to analyze discovered dependence.

\subsection{FourRoom, LavaGap and DoorKey}
In these task, the agent needs to navigate in a maze  and a reward would be obtained when reaching a specific target position~\cite{gym_minigrid}. FourRoom cantains a four-room maze interconnected by 4 corridors (gaps in the walls), LavaGap has  deadly lava distributed in the room, and in DoorKey the agent must first pick up a key before unlocking the door to reach the target square in the other room.  There are four actions:  clockwise rotation, anticlockwise rotation, moving forward, and pick-up. In total, each dataset contains $4,000$ sequences 

%The action space contains three concrete actions: clockwise rotation, anticlockwise rotation, and moving forward. Each action taken has a cost, and the episode (a round of random running) would be terminated when arriving at the target or when the accumulated cost exceeds the reward. Demonstrations are collected from a pre-trained agent. In total, the dataset contains $4,000$ sequences (average trajectory length: $22.17$).
%\subsection{LavaGap}
%This environment requires the agent to navigate in a one-room maze to reach the target region. The target region is put at the opposite corner, and there are deadly lava distributed in the room. Touching the lava would immediately terminate the current episode, and the agent must navigate through a narrow gap to safely arrive at the target. Similar to the FourRoom dataset, there are three candidate actions, and we obtain $4,000$ sequences of average length $6.12$ with a pre-trained neural agent. 
%\subsection{Doorkey}
%This environment is a maze composed of two rooms, in which the agent must first pick up a key before unlocking the door to reach the target square in the other room. There are four actions:  clockwise rotation, anticlockwise rotation, moving forward, and pick-up. using a pre-trained agent, $4,000$ sequences of average length $9.83$ are collected to construct this dataset.

\begin{figure}[t]
  \centering
  \subfigure[Kura10,Kura10\_vary]{
		\includegraphics[width=0.23\textwidth]{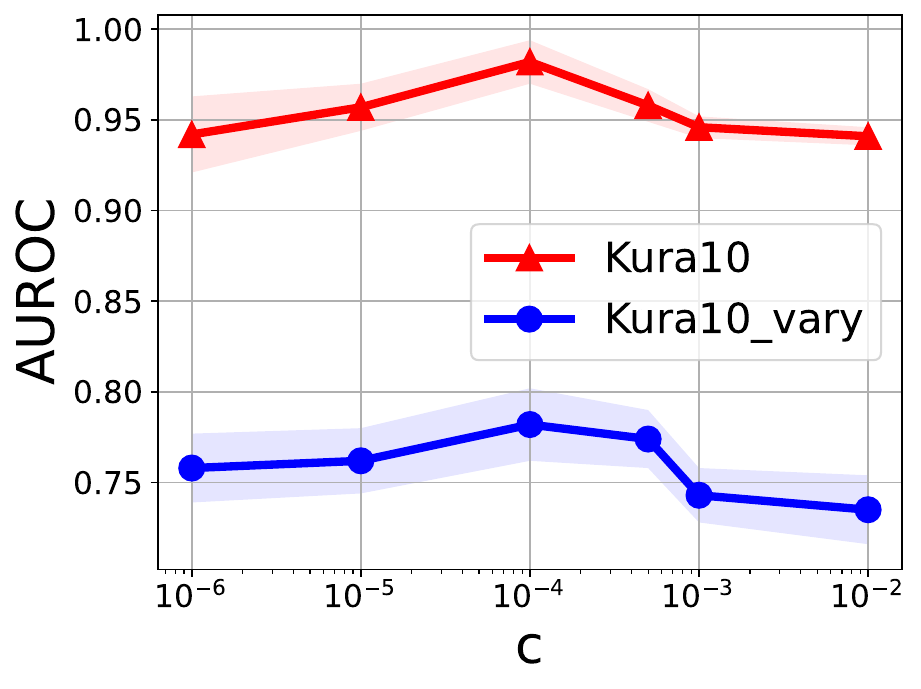}}
  \subfigure[LavaGap, FourRoom]{
		\includegraphics[width=0.23\textwidth]{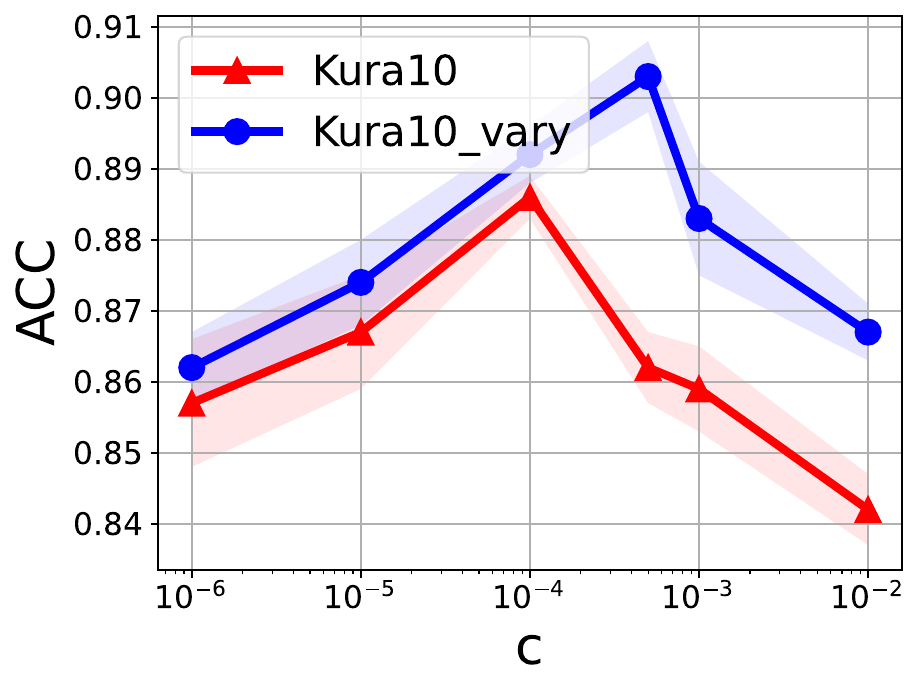}}
    \vskip -1.5em
    \caption{Sensitivity on $c$, i.e, acyclicity regularization $\mathcal{R}_{DAG}$ } 
		\label{fig:acyc}
    \vskip -1.5em
\end{figure}

\section{Acyclicity Regularization}\label{ap:sensitivity}
In this part, we analyze the sensitivity of {\method} on the importance of acyclicity regularization. $c$ is the penalty parameter in the Augmented Lagrangian algorithm and controls the weight of $\mathcal{R}_{DAG}$, as defined in Eq.~\ref{eq:targetAll}. Acyclicity constraint discourages the existence of loops in discovered causal graphs. We vary $c$ as $\{10^{-6}, 10^{-5},\dots,10^{-2}\}$. Again, experiments are evaluated w.r.t both causal discovery and action prediction. Each experiment is conducted $3$ times and the average results in causal discovery are shown in Fig~\ref{fig:acyc}. We can observe that the model performs best with $c \in [1e-5,5e-4]$. When $c$ is set to a high value, the importance of acyclicity regularization will increase quickly in the early stage, which could be misleading.

\begin{figure*}[t]
  \centering
  \subfigure[Stage $1$]{
		\includegraphics[width=0.24\textwidth]{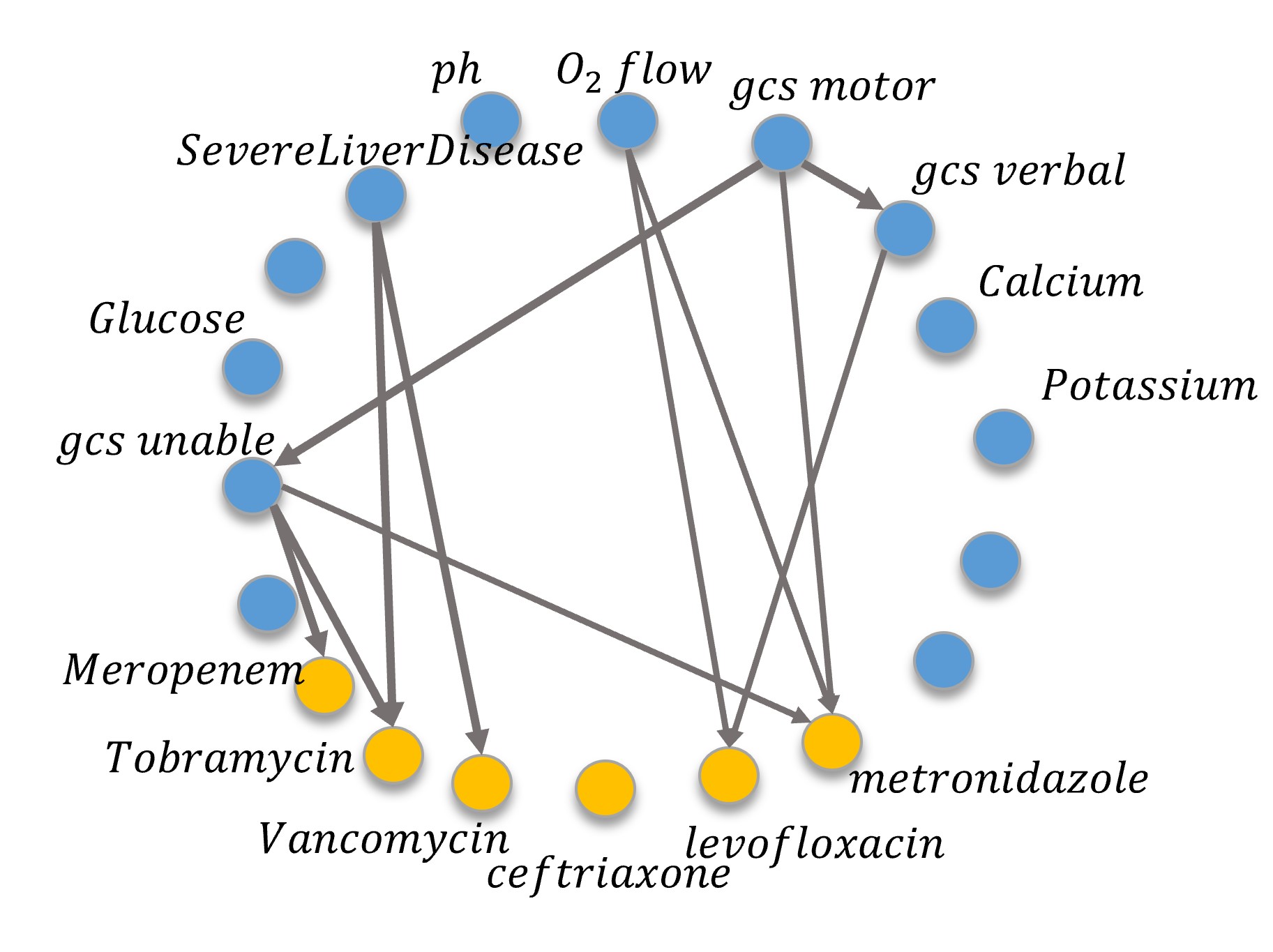}} ~~~
  \subfigure[Stage $2$]{
		\includegraphics[width=0.24\textwidth]{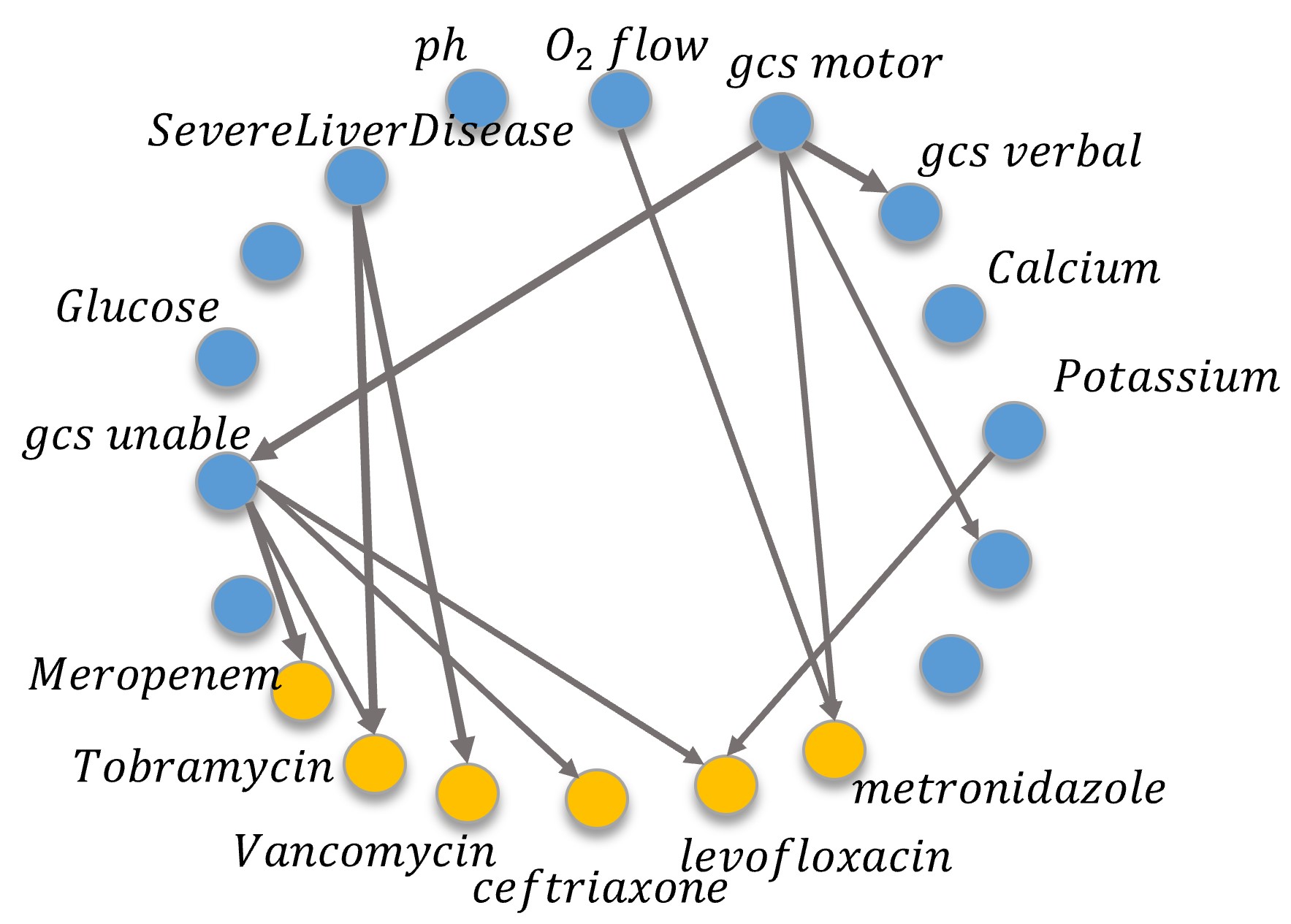}} ~~~
  \subfigure[Stage $3$]{
		\includegraphics[width=0.24\textwidth]{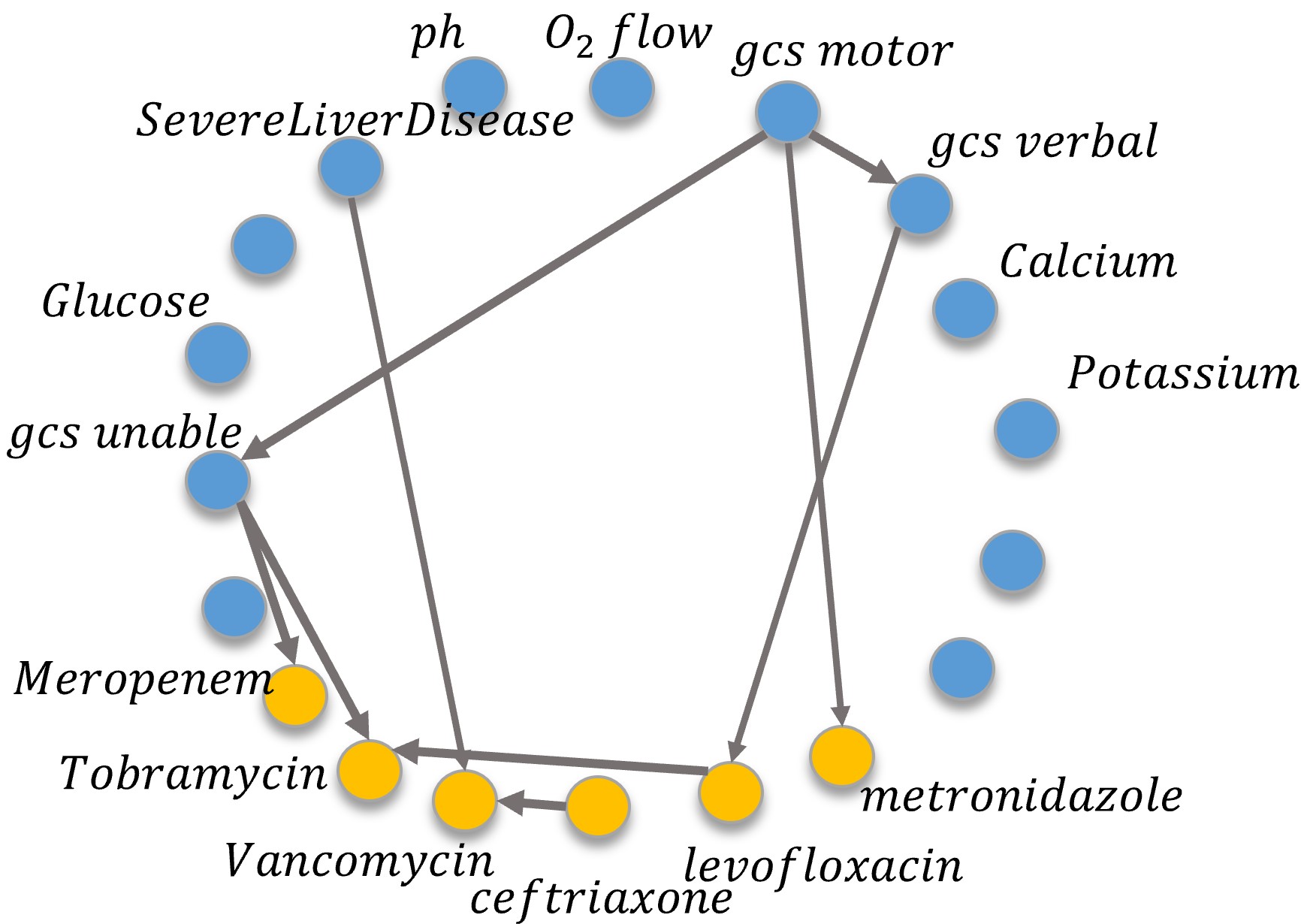}}
    \vskip -1.5em
    \caption{Discovered DAG templates on Mimic-IV Dataset. For ease of analysis, we visualize only a subset of states (blue nodes) and actions (yellow nodes).} 
		\label{fig:templates}
    \vskip -1.5em
\end{figure*}

\section{Case Study - Mimic}\label{ap:case1}
In this section, we conduct a case study on MIMIC-IV to present examples of identified causal relations. Concretely, patients diagnosed with comorbidity are used, and $M$ is set to $3$ reflecting the severity of sepsis~\cite{buras2005animal}. For simplicity in analysis, we select only a subset of nodes, and learned DAG templates are visualized in Figure~\ref{fig:templates}, with edge width denoting importance weight. For ease of interpretation, only top edges are drawn. With knowledge from domain experts (doctors), we learn that Meropenem, Tobramycin, and Vancomycin are typically used for severe sepsis, while ceftriaxone, levofloxacin, and metronidazole are for mild sepsis. From the figure, it is shown that:
\begin{itemize}[leftmargin=*]
    \item Across templates, strong causal relations can be observed from ``SevereLiverDesease'' to Meropenem and Tobramycin, or from ``gcs\_unable'' to Vancomycin. It is in accordance with prior knowledge as these symptoms denote severe health conditions; 
    \item With sepsis becoming more severe, as in the third template, previous usage of mild sepsis drugs has a causal effect on severe sepsis drugs. It obeys the rule observed in ~\cite{li2020temporal} that mild drugs must be used in advance due to antibiotic resistance.
    \item {\method} captures strong conditional dependence of ``gcs\_unable'' (measuring consciousness of patients) and ``gcs\_verbal'' (ability to speak) on ``gcs\_motor'' (normal working state of muscle), which is reasonable and easy to interpret.
\end{itemize}
These observations verify the ability of proposed approach in exposing causal relations and can increase the interpretability.

\section{Case Study - Kura}\label{ap:case}

\begin{figure}[t]
  \centering
  \subfigure[GT]{
		\includegraphics[width=0.11\textwidth]{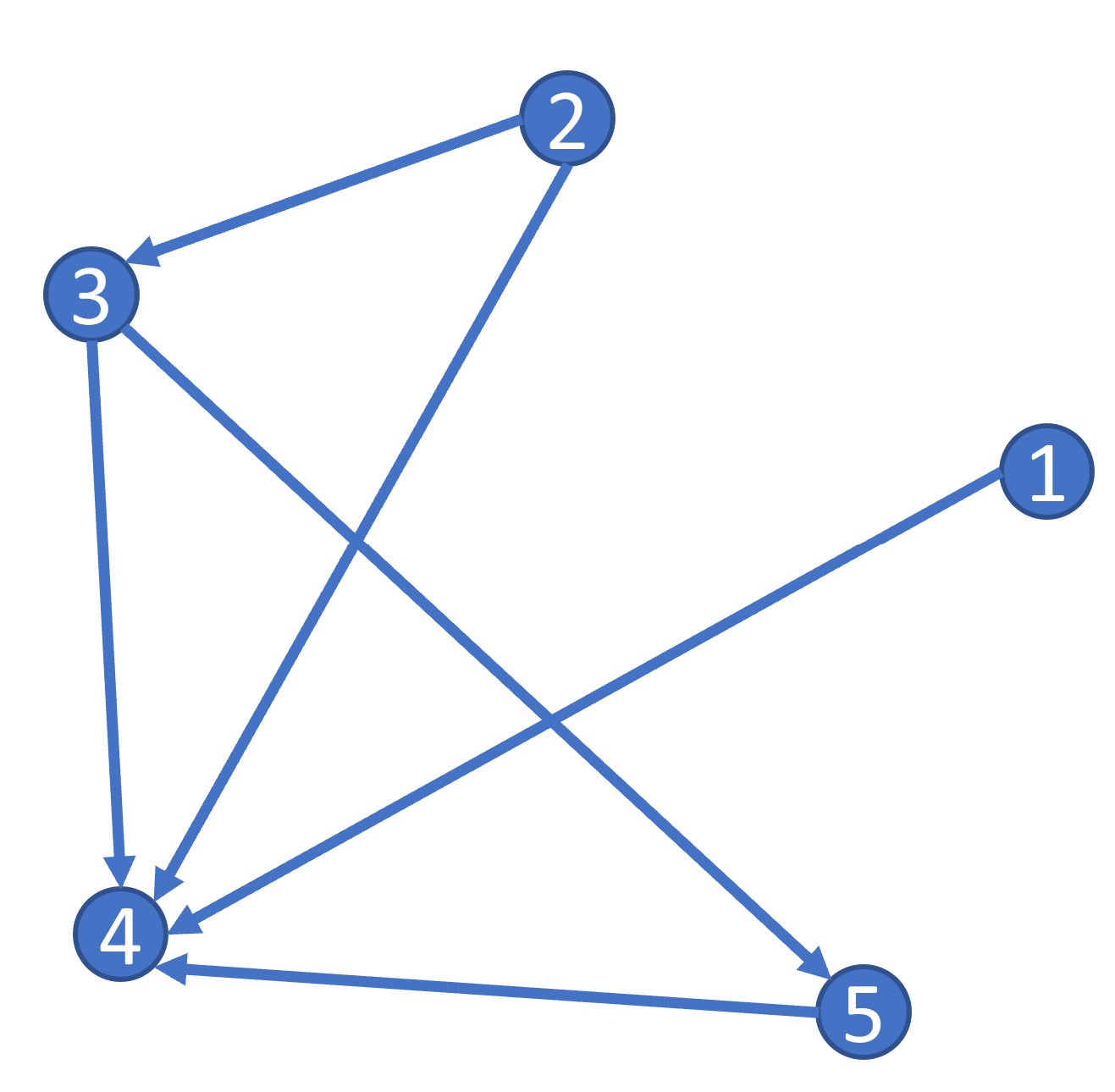}}
  \subfigure[Template]{
		\includegraphics[width=0.11\textwidth]{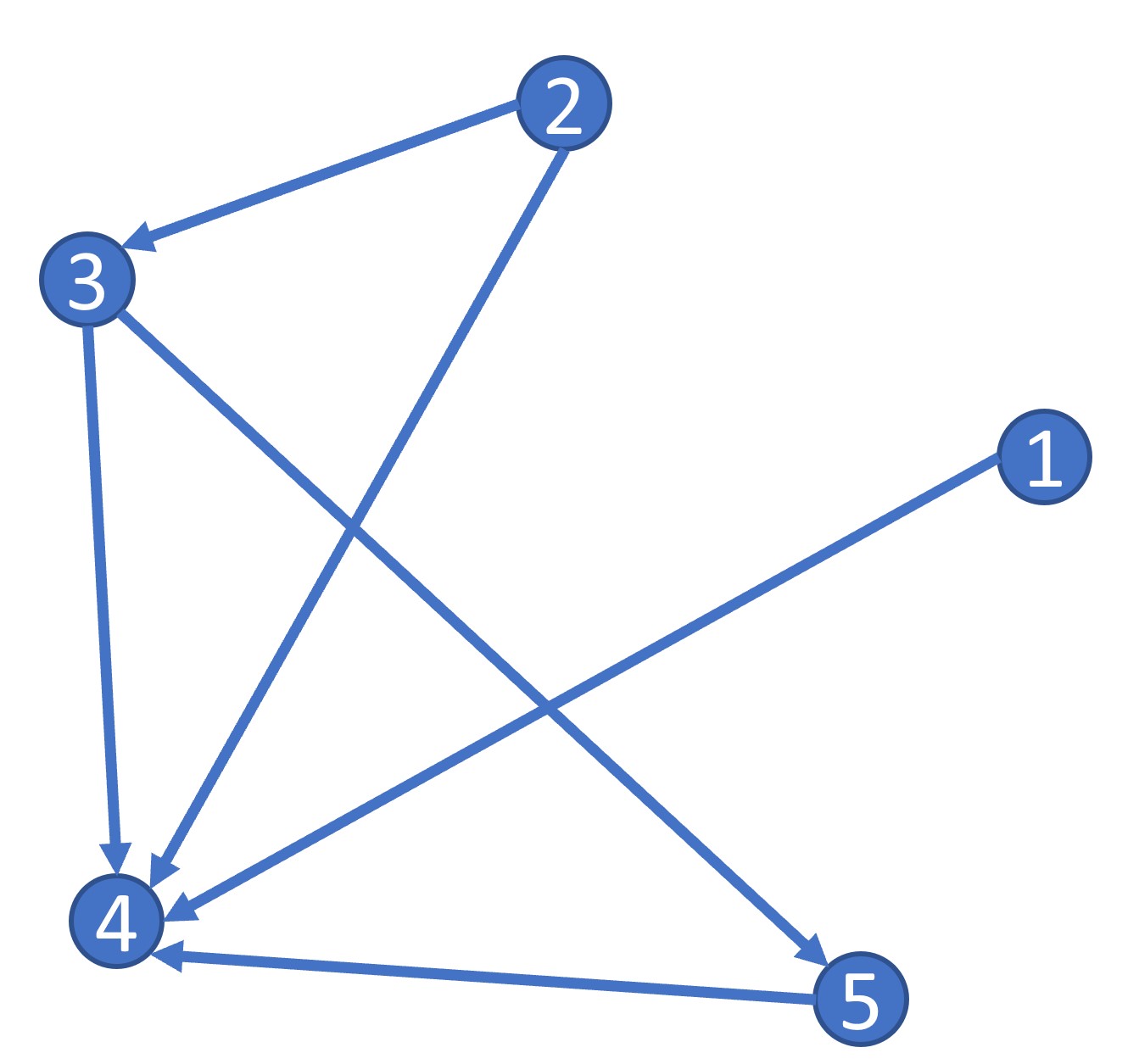}}
  \subfigure[GT]{
		\includegraphics[width=0.11\textwidth]{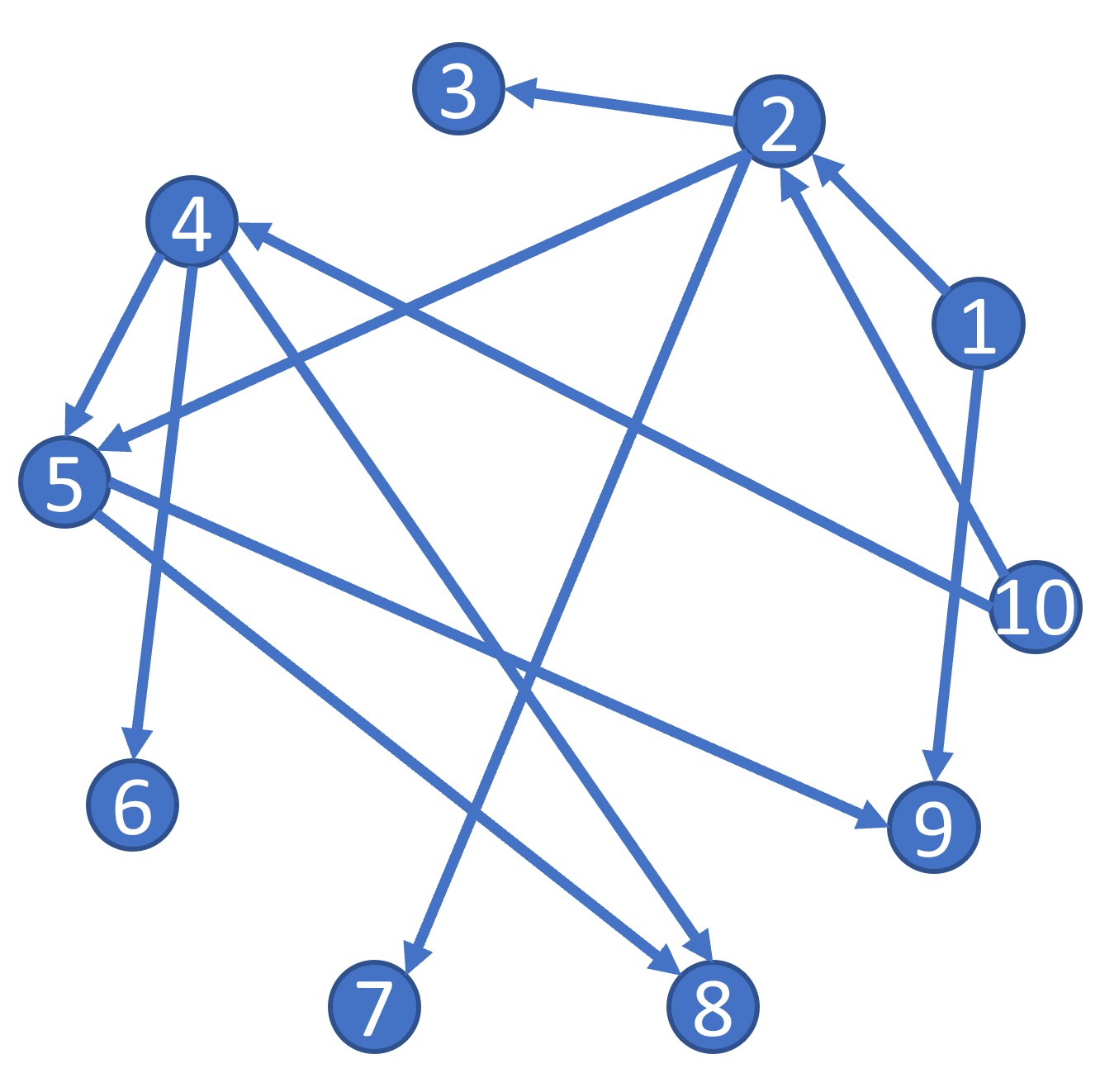}}
  \subfigure[Template]{
		\includegraphics[width=0.11\textwidth]{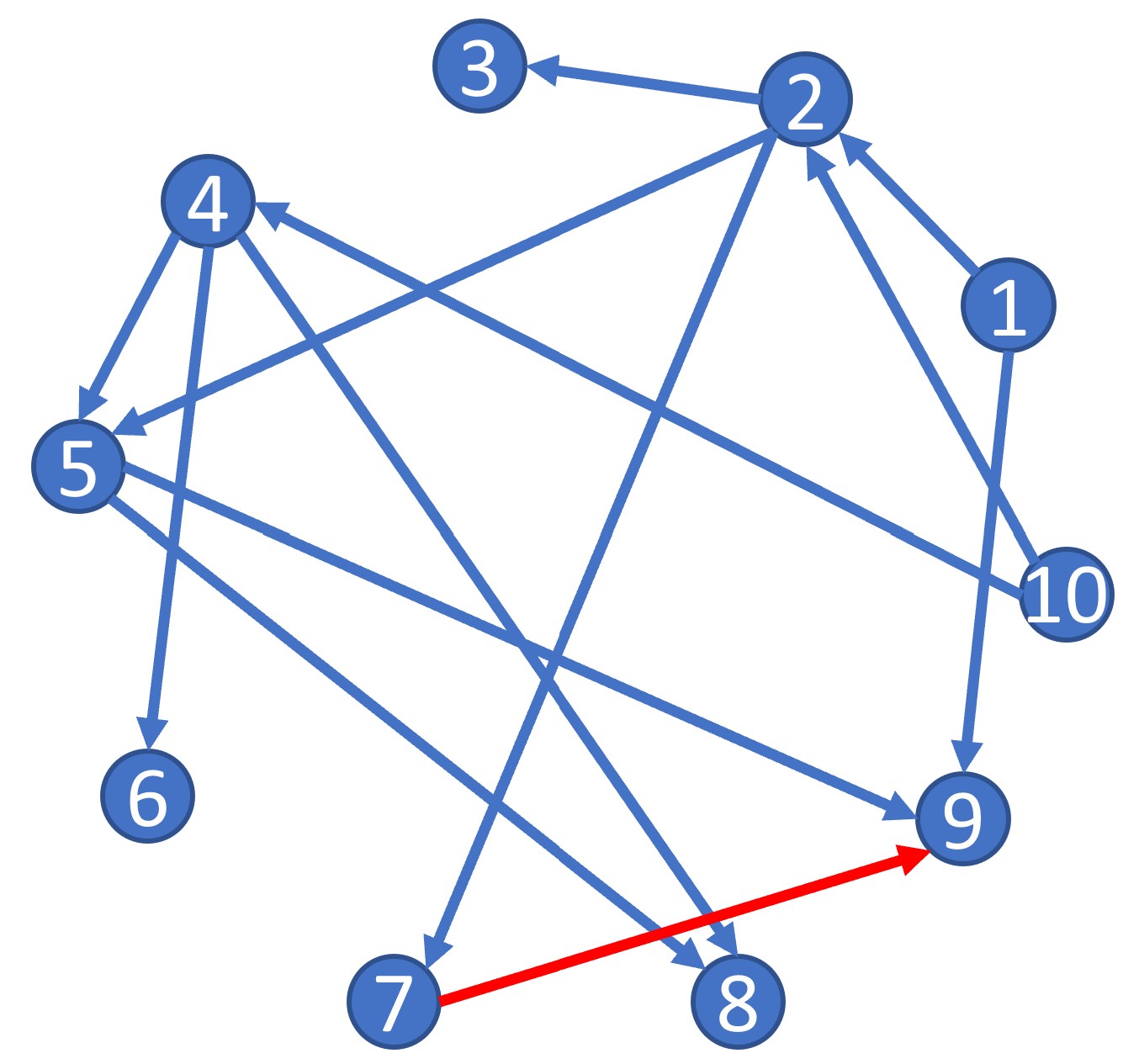}}
		
    \caption{Case Study on Kura5 and Kura10 respectively. } 
		\label{fig:static_example}
    \vskip -1em
\end{figure}
\begin{figure}[t]
  \centering
  \subfigure[GT DAG $1$ ]{
		\includegraphics[width=0.13\textwidth]{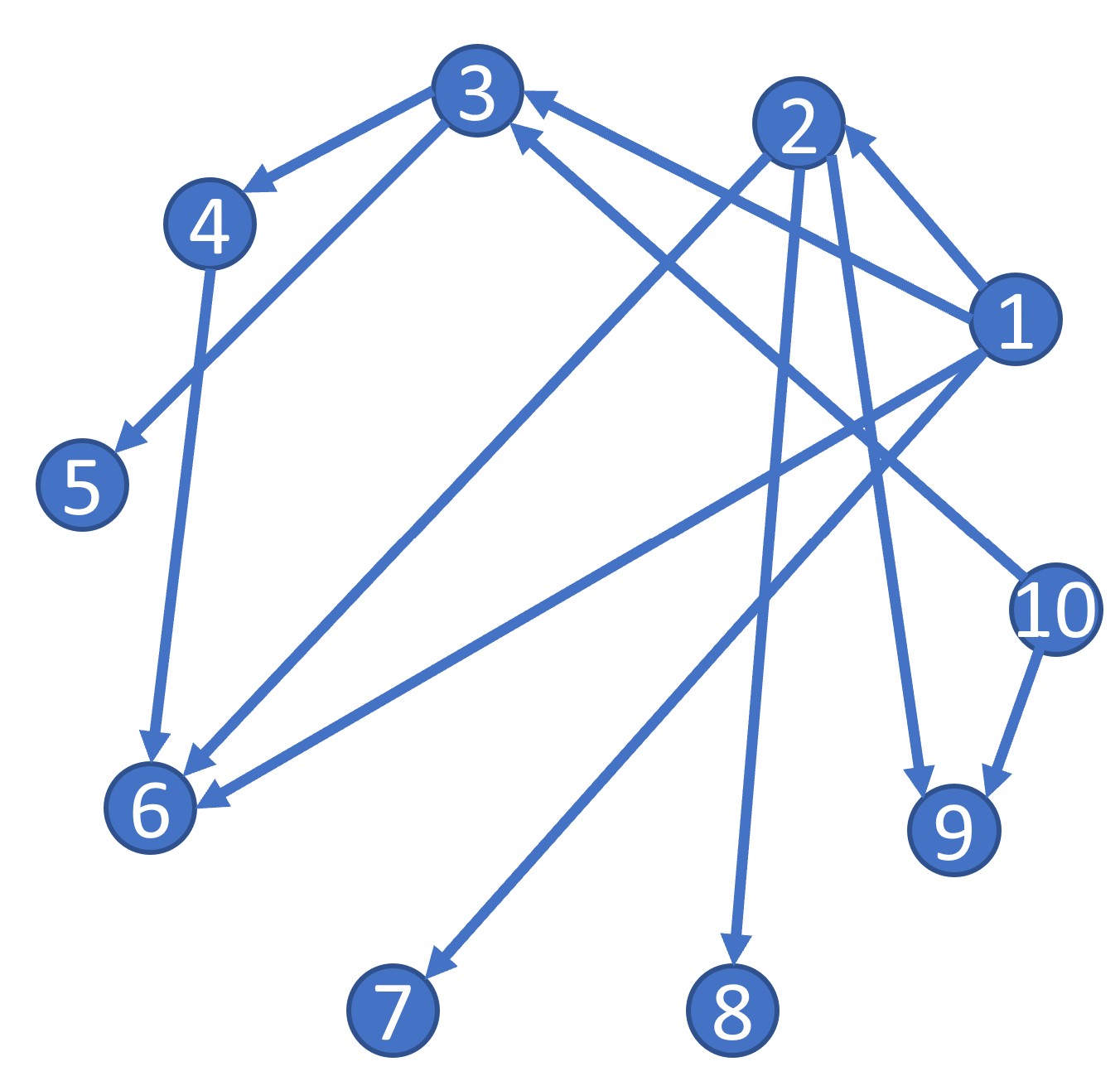}}
  \subfigure[GT DAG $2$ ]{
		\includegraphics[width=0.13\textwidth]{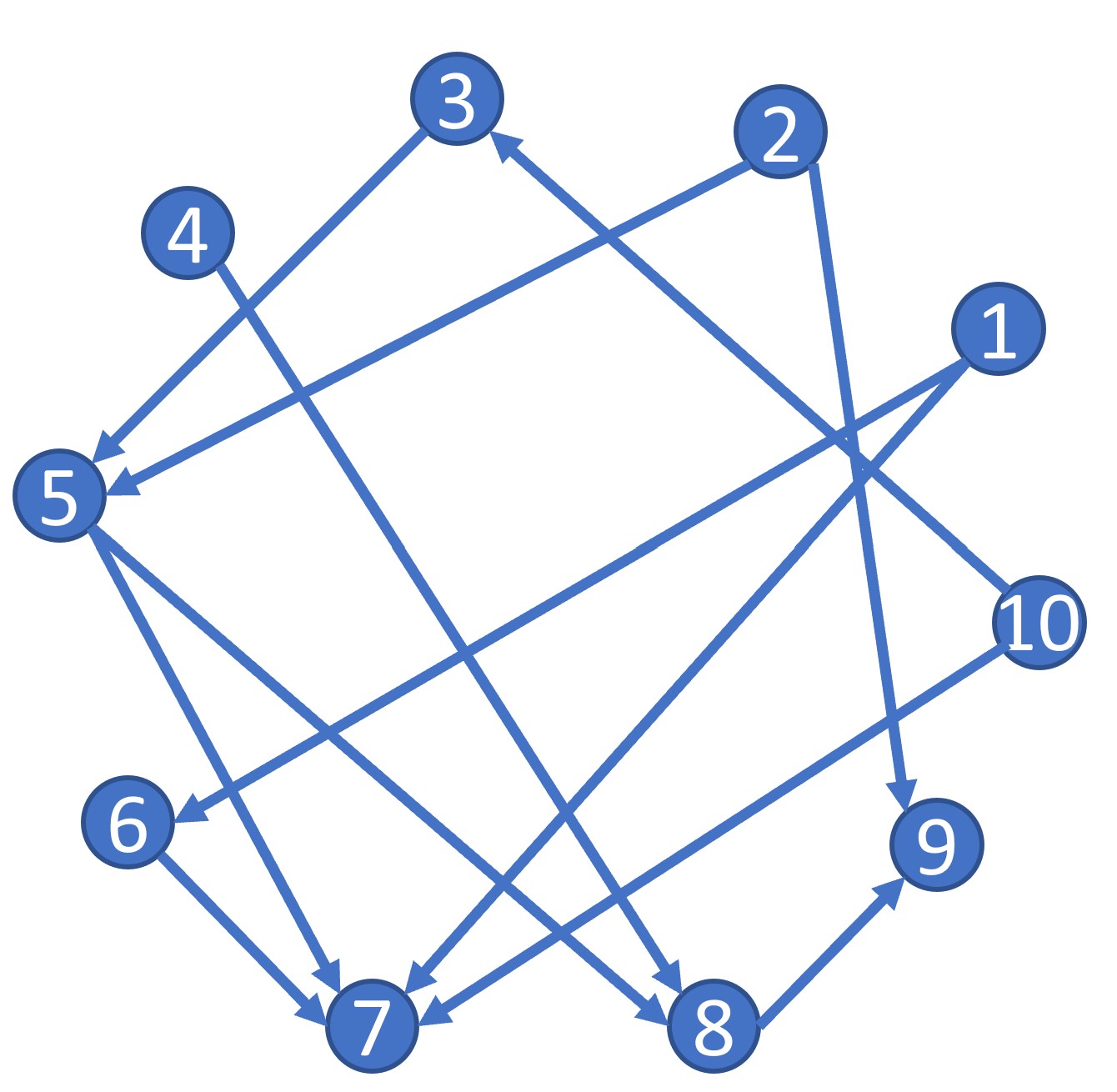}}
  \subfigure[GT DAG $3$ ]{
		\includegraphics[width=0.13\textwidth]{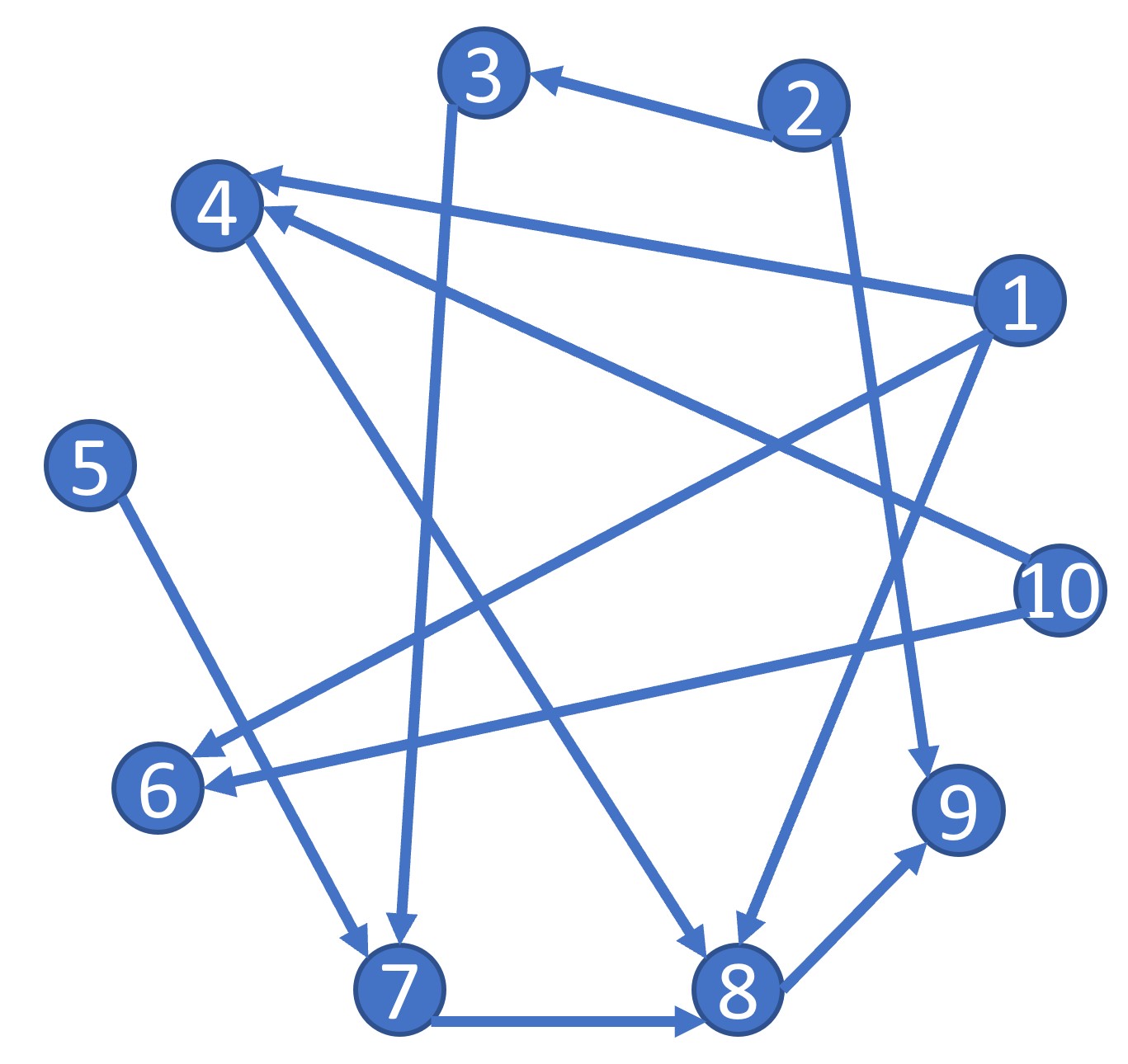}}\\
  \subfigure[Template$1$ ]{
		\includegraphics[width=0.13\textwidth]{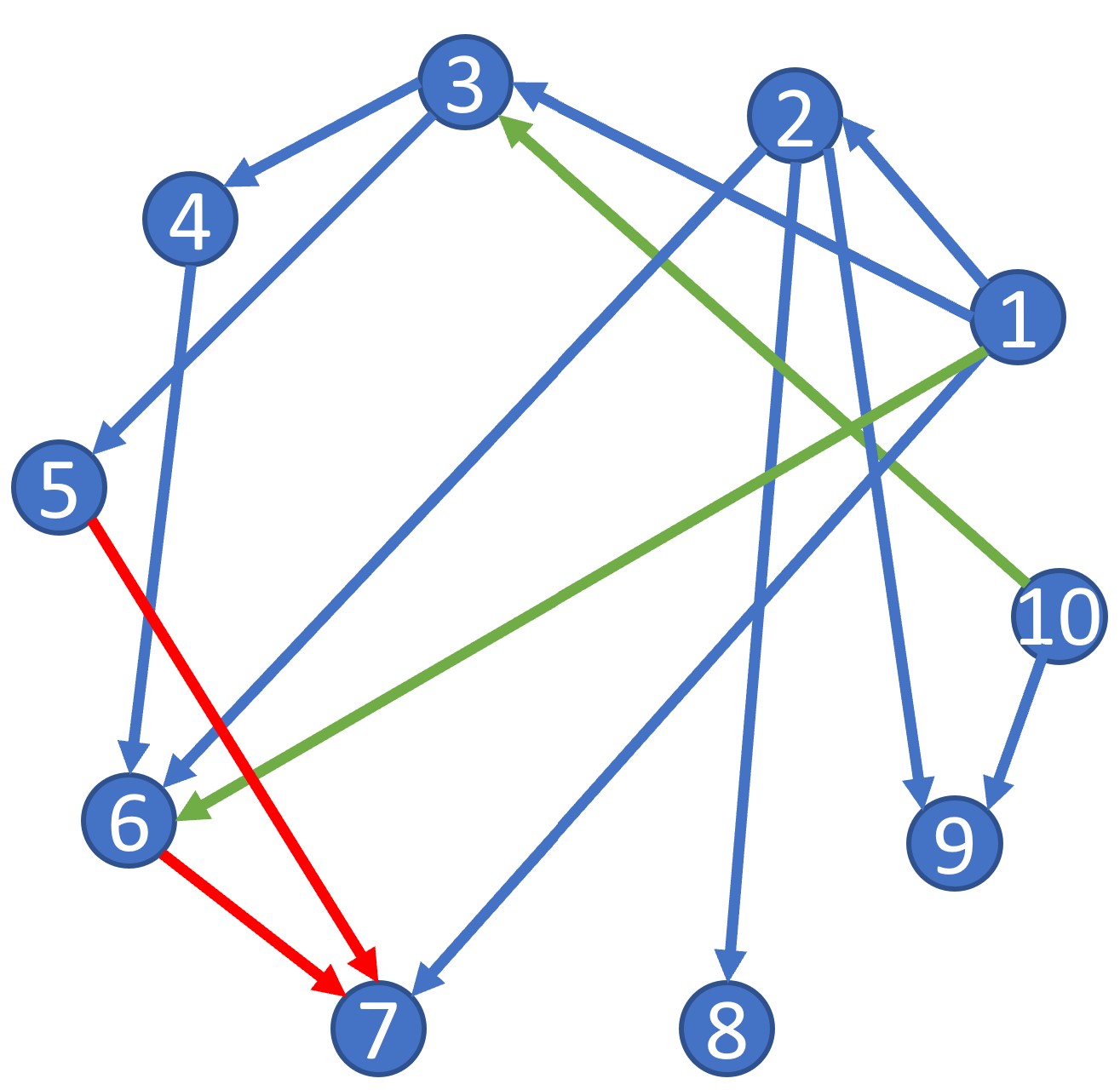}}
  \subfigure[Template$2$ ]{
		\includegraphics[width=0.13\textwidth]{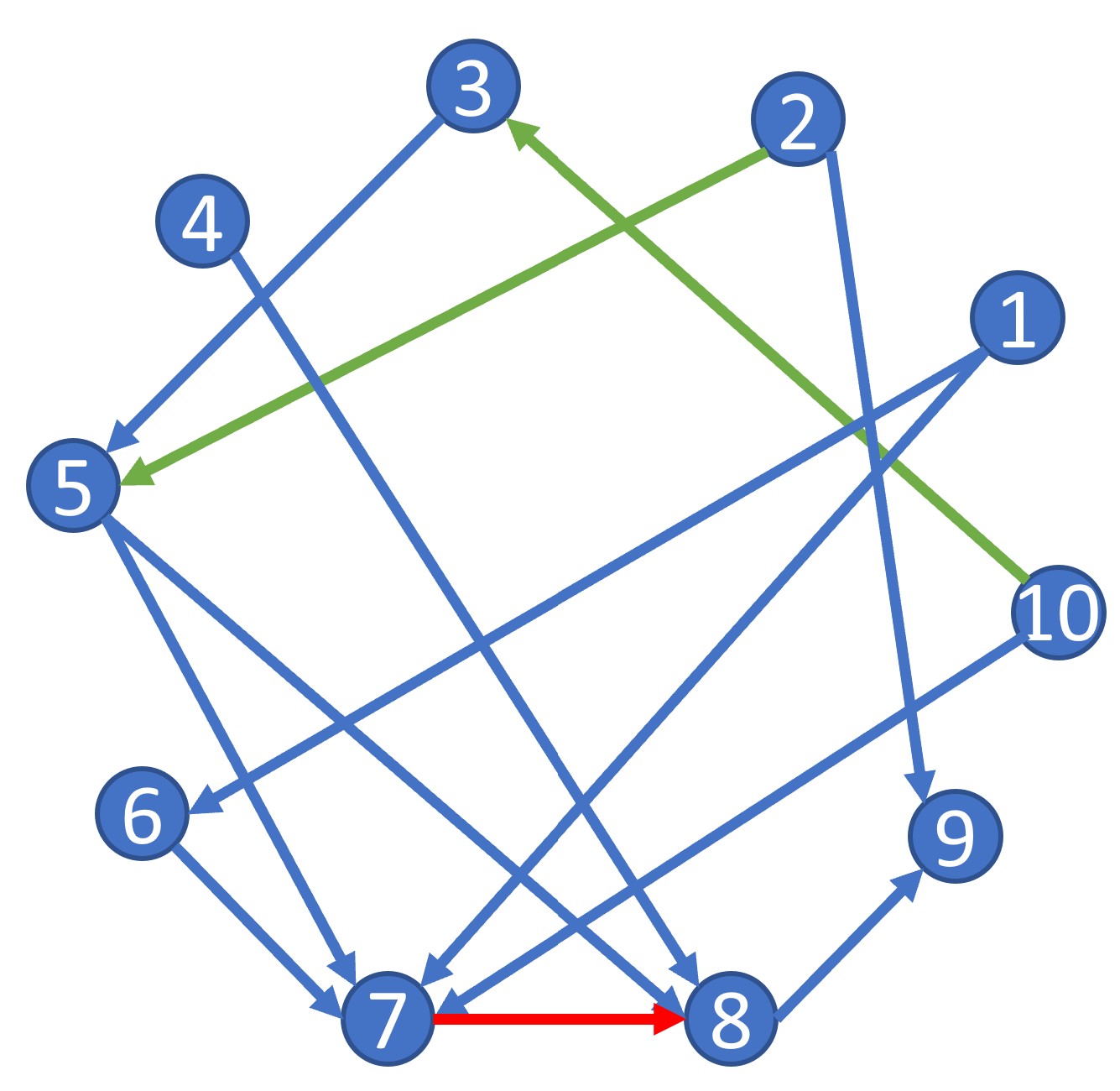}}
  \subfigure[Template$3$ ]{
		\includegraphics[width=0.13\textwidth]{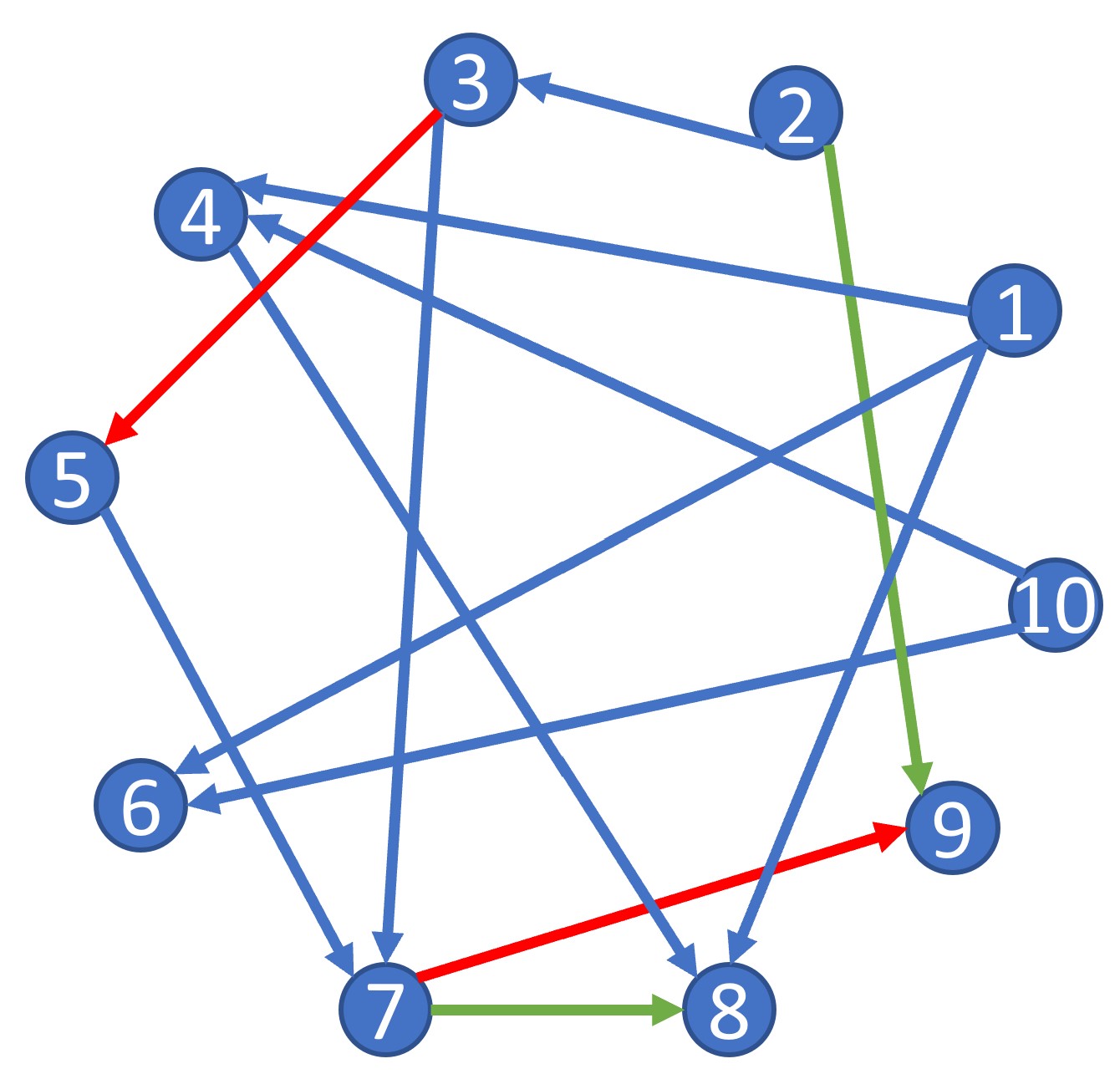}}
		
    \caption{Case Study on Kura10\_vary. } 
		\label{fig:dynamic_example}
    \vskip -1em
\end{figure}

In this section, we include more examples to show the causal graphs discovered by our approach. On the simulation dataset Kuramoto, ground-truth DAGs are available, making it easier to examine the correctness of identified causal edges. Specifically, examples are provided both for the static case and dynamic case, with examples showing in Figure~\ref{fig:static_example} and Figure~\ref{fig:dynamic_example} respectively. In Figure~\ref{fig:static_example}, learned causal graphs on Kura5 and Kura10 are compared with the ground-truths respectively. Edge width represents learned edge weight, and red edges denote erroneous causal edges. It can be observed that discovered causal graph aligns well with the oracle DAG well, demonstrating quality of explanation provided by {\method}. In Figure~\ref{fig:dynamic_example}, we further conduct a case study in the dynamic setting, on Kura10\_vary. Red edges denote erroneous causal edges while green edges denote missing causal edges. This dynamic causal discovery task is more challenging than the static version, and the number of erroneous or missing edges would increase. Generally, results show that {\method} can work well in this setting, successfully exposing most causal edges. These results further show the ability of {\method} in conducting self-explainable imitation learning from the perspective of causal discovery, providing its captured variable relations.

%% file: main.bbl
%%% -*-BibTeX-*-
%%% Do NOT edit. File created by BibTeX with style
%%% ACM-Reference-Format-Journals [18-Jan-2012].

\begin{thebibliography}{51}

%%% ====================================================================
%%% NOTE TO THE USER: you can override these defaults by providing
%%% customized versions of any of these macros before the \bibliography
%%% command.  Each of them MUST provide its own final punctuation,
%%% except for \shownote{}, \showDOI{}, and \showURL{}.  The latter two
%%% do not use final punctuation, in order to avoid confusing it with
%%% the Web address.
%%%
%%% To suppress output of a particular field, define its macro to expand
%%% to an empty string, or better, \unskip, like this:
%%%
%%% \newcommand{\showDOI}[1]{\unskip}   % LaTeX syntax
%%%
%%% \def \showDOI #1{\unskip}           % plain TeX syntax
%%%
%%% ====================================================================

\ifx \showCODEN    \undefined \def \showCODEN     #1{\unskip}     \fi
\ifx \showDOI      \undefined \def \showDOI       #1{#1}\fi
\ifx \showISBNx    \undefined \def \showISBNx     #1{\unskip}     \fi
\ifx \showISBNxiii \undefined \def \showISBNxiii  #1{\unskip}     \fi
\ifx \showISSN     \undefined \def \showISSN      #1{\unskip}     \fi
\ifx \showLCCN     \undefined \def \showLCCN      #1{\unskip}     \fi
\ifx \shownote     \undefined \def \shownote      #1{#1}          \fi
\ifx \showarticletitle \undefined \def \showarticletitle #1{#1}   \fi
\ifx \showURL      \undefined \def \showURL       {\relax}        \fi
% The following commands are used for tagged output and should be
% invisible to TeX
\providecommand\bibfield[2]{#2}
\providecommand\bibinfo[2]{#2}
\providecommand\natexlab[1]{#1}
\providecommand\showeprint[2][]{arXiv:#2}

\bibitem[\protect\citeauthoryear{Abbeel, Coates, Quigley, and Ng}{Abbeel
  et~al\mbox{.}}{2007}]%
        {abbeel2007application}
\bibfield{author}{\bibinfo{person}{Pieter Abbeel}, \bibinfo{person}{Adam
  Coates}, \bibinfo{person}{Morgan Quigley}, {and} \bibinfo{person}{Andrew~Y
  Ng}.} \bibinfo{year}{2007}\natexlab{}.
\newblock \showarticletitle{An application of reinforcement learning to
  aerobatic helicopter flight}.
\newblock \bibinfo{journal}{\emph{Advances in neural information processing
  systems}}  \bibinfo{volume}{19} (\bibinfo{year}{2007}), \bibinfo{pages}{1}.
\newblock


\bibitem[\protect\citeauthoryear{Balcilar, Van~Eyden, Uwilingiye, and
  Gupta}{Balcilar et~al\mbox{.}}{2017}]%
        {balcilar2017impact}
\bibfield{author}{\bibinfo{person}{Mehmet Balcilar}, \bibinfo{person}{Rene{\'e}
  Van~Eyden}, \bibinfo{person}{Josine Uwilingiye}, {and}
  \bibinfo{person}{Rangan Gupta}.} \bibinfo{year}{2017}\natexlab{}.
\newblock \showarticletitle{The impact of oil price on South African GDP
  growth: A Bayesian Markov switching-VAR analysis}.
\newblock \bibinfo{journal}{\emph{African Development Review}}
  \bibinfo{volume}{29}, \bibinfo{number}{2} (\bibinfo{year}{2017}),
  \bibinfo{pages}{319--336}.
\newblock


\bibitem[\protect\citeauthoryear{Bressler and Seth}{Bressler and Seth}{2011}]%
        {bressler2011wiener}
\bibfield{author}{\bibinfo{person}{Steven~L Bressler} {and}
  \bibinfo{person}{Anil~K Seth}.} \bibinfo{year}{2011}\natexlab{}.
\newblock \showarticletitle{Wiener--Granger causality: a well established
  methodology}.
\newblock \bibinfo{journal}{\emph{Neuroimage}} \bibinfo{volume}{58},
  \bibinfo{number}{2} (\bibinfo{year}{2011}), \bibinfo{pages}{323--329}.
\newblock


\bibitem[\protect\citeauthoryear{Buras, Holzmann, and Sitkovsky}{Buras
  et~al\mbox{.}}{2005}]%
        {buras2005animal}
\bibfield{author}{\bibinfo{person}{Jon~A Buras}, \bibinfo{person}{Bernhard
  Holzmann}, {and} \bibinfo{person}{Michail Sitkovsky}.}
  \bibinfo{year}{2005}\natexlab{}.
\newblock \showarticletitle{Animal models of sepsis: setting the stage}.
\newblock \bibinfo{journal}{\emph{Nature reviews Drug discovery}}
  \bibinfo{volume}{4}, \bibinfo{number}{10} (\bibinfo{year}{2005}),
  \bibinfo{pages}{854--865}.
\newblock


\bibitem[\protect\citeauthoryear{Chevalier-Boisvert, Willems, and
  Pal}{Chevalier-Boisvert et~al\mbox{.}}{2018}]%
        {gym_minigrid}
\bibfield{author}{\bibinfo{person}{Maxime Chevalier-Boisvert},
  \bibinfo{person}{Lucas Willems}, {and} \bibinfo{person}{Suman Pal}.}
  \bibinfo{year}{2018}\natexlab{}.
\newblock \bibinfo{title}{Minimalistic Gridworld Environment for OpenAI Gym}.
\newblock
  \bibinfo{howpublished}{\url{https://github.com/maximecb/gym-minigrid}}.
\newblock


\bibitem[\protect\citeauthoryear{Codevilla, M{\"u}ller, L{\'o}pez, Koltun, and
  Dosovitskiy}{Codevilla et~al\mbox{.}}{2018}]%
        {codevilla2018end}
\bibfield{author}{\bibinfo{person}{Felipe Codevilla}, \bibinfo{person}{Matthias
  M{\"u}ller}, \bibinfo{person}{Antonio L{\'o}pez}, \bibinfo{person}{Vladlen
  Koltun}, {and} \bibinfo{person}{Alexey Dosovitskiy}.}
  \bibinfo{year}{2018}\natexlab{}.
\newblock \showarticletitle{End-to-end driving via conditional imitation
  learning}. In \bibinfo{booktitle}{\emph{2018 IEEE International Conference on
  Robotics and Automation (ICRA)}}. IEEE, \bibinfo{pages}{4693--4700}.
\newblock


\bibitem[\protect\citeauthoryear{de~Haan, Jayaraman, and Levine}{de~Haan
  et~al\mbox{.}}{2019}]%
        {de2019causal}
\bibfield{author}{\bibinfo{person}{Pim de Haan}, \bibinfo{person}{Dinesh
  Jayaraman}, {and} \bibinfo{person}{Sergey Levine}.}
  \bibinfo{year}{2019}\natexlab{}.
\newblock \showarticletitle{Causal confusion in imitation learning}.
\newblock \bibinfo{journal}{\emph{Advances in Neural Information Processing
  Systems}}  \bibinfo{volume}{32} (\bibinfo{year}{2019}),
  \bibinfo{pages}{11698--11709}.
\newblock


\bibitem[\protect\citeauthoryear{Frausto, Pittet, Hwang, Woolson, and
  Wenzel}{Frausto et~al\mbox{.}}{1998}]%
        {frausto1998dynamics}
\bibfield{author}{\bibinfo{person}{M~Sigfrido~Rangel Frausto},
  \bibinfo{person}{Didier Pittet}, \bibinfo{person}{Taekyu Hwang},
  \bibinfo{person}{Robert~F Woolson}, {and} \bibinfo{person}{Richard~P
  Wenzel}.} \bibinfo{year}{1998}\natexlab{}.
\newblock \showarticletitle{The dynamics of disease progression in sepsis:
  Markov modeling describing the natural history and the likely impact of
  effective antisepsis agents}.
\newblock \bibinfo{journal}{\emph{Clinical infectious diseases}}
  \bibinfo{volume}{27}, \bibinfo{number}{1} (\bibinfo{year}{1998}),
  \bibinfo{pages}{185--190}.
\newblock


\bibitem[\protect\citeauthoryear{Fu, Luo, and Levine}{Fu et~al\mbox{.}}{2018}]%
        {fu2018learning}
\bibfield{author}{\bibinfo{person}{Justin Fu}, \bibinfo{person}{Katie Luo},
  {and} \bibinfo{person}{Sergey Levine}.} \bibinfo{year}{2018}\natexlab{}.
\newblock \showarticletitle{Learning Robust Rewards with Adverserial Inverse
  Reinforcement Learning}. In \bibinfo{booktitle}{\emph{International
  Conference on Learning Representations}}.
\newblock


\bibitem[\protect\citeauthoryear{Granger}{Granger}{1969}]%
        {granger1969investigating}
\bibfield{author}{\bibinfo{person}{Clive~WJ Granger}.}
  \bibinfo{year}{1969}\natexlab{}.
\newblock \showarticletitle{Investigating causal relations by econometric
  models and cross-spectral methods}.
\newblock \bibinfo{journal}{\emph{Econometrica: journal of the Econometric
  Society}} (\bibinfo{year}{1969}), \bibinfo{pages}{424--438}.
\newblock


\bibitem[\protect\citeauthoryear{Gu, Holly, Lillicrap, and Levine}{Gu
  et~al\mbox{.}}{2017}]%
        {gu2017deep}
\bibfield{author}{\bibinfo{person}{Shixiang Gu}, \bibinfo{person}{Ethan Holly},
  \bibinfo{person}{Timothy Lillicrap}, {and} \bibinfo{person}{Sergey Levine}.}
  \bibinfo{year}{2017}\natexlab{}.
\newblock \showarticletitle{Deep reinforcement learning for robotic
  manipulation with asynchronous off-policy updates}. In
  \bibinfo{booktitle}{\emph{2017 IEEE international conference on robotics and
  automation (ICRA)}}. IEEE, \bibinfo{pages}{3389--3396}.
\newblock


\bibitem[\protect\citeauthoryear{Hallac, Park, Boyd, and Leskovec}{Hallac
  et~al\mbox{.}}{2017}]%
        {hallac2017network}
\bibfield{author}{\bibinfo{person}{David Hallac}, \bibinfo{person}{Youngsuk
  Park}, \bibinfo{person}{Stephen Boyd}, {and} \bibinfo{person}{Jure
  Leskovec}.} \bibinfo{year}{2017}\natexlab{}.
\newblock \showarticletitle{Network inference via the time-varying graphical
  lasso}. In \bibinfo{booktitle}{\emph{Proceedings of the 23rd ACM SIGKDD
  International Conference on Knowledge Discovery and Data Mining}}.
  \bibinfo{pages}{205--213}.
\newblock


\bibitem[\protect\citeauthoryear{Hamilton, Ying, and Leskovec}{Hamilton
  et~al\mbox{.}}{2017}]%
        {Hamilton2017InductiveRL}
\bibfield{author}{\bibinfo{person}{William~L. Hamilton},
  \bibinfo{person}{Zhitao Ying}, {and} \bibinfo{person}{J. Leskovec}.}
  \bibinfo{year}{2017}\natexlab{}.
\newblock \showarticletitle{Inductive Representation Learning on Large Graphs}.
  In \bibinfo{booktitle}{\emph{NIPS}}.
\newblock


\bibitem[\protect\citeauthoryear{Haufe, M{\"u}ller, Nolte, and
  Kr{\"a}mer}{Haufe et~al\mbox{.}}{2010}]%
        {haufe2010sparse}
\bibfield{author}{\bibinfo{person}{Stefan Haufe}, \bibinfo{person}{Klaus-Robert
  M{\"u}ller}, \bibinfo{person}{Guido Nolte}, {and} \bibinfo{person}{Nicole
  Kr{\"a}mer}.} \bibinfo{year}{2010}\natexlab{}.
\newblock \showarticletitle{Sparse causal discovery in multivariate time
  series}. In \bibinfo{booktitle}{\emph{Causality: Objectives and Assessment}}.
  PMLR, \bibinfo{pages}{97--106}.
\newblock


\bibitem[\protect\citeauthoryear{Ho and Ermon}{Ho and Ermon}{2016}]%
        {ho2016generative}
\bibfield{author}{\bibinfo{person}{Jonathan Ho} {and} \bibinfo{person}{Stefano
  Ermon}.} \bibinfo{year}{2016}\natexlab{}.
\newblock \showarticletitle{Generative adversarial imitation learning}.
\newblock \bibinfo{journal}{\emph{Advances in neural information processing
  systems}}  \bibinfo{volume}{29} (\bibinfo{year}{2016}),
  \bibinfo{pages}{4565--4573}.
\newblock


\bibitem[\protect\citeauthoryear{Hsiao and Lozano-Perez}{Hsiao and
  Lozano-Perez}{2006}]%
        {hsiao2006imitation}
\bibfield{author}{\bibinfo{person}{Kaijen Hsiao} {and} \bibinfo{person}{Tomas
  Lozano-Perez}.} \bibinfo{year}{2006}\natexlab{}.
\newblock \showarticletitle{Imitation learning of whole-body grasps}. In
  \bibinfo{booktitle}{\emph{2006 IEEE/RSJ international conference on
  intelligent robots and systems}}. IEEE, \bibinfo{pages}{5657--5662}.
\newblock


\bibitem[\protect\citeauthoryear{Hsieh, Sun, Tang, Wang, and Honavar}{Hsieh
  et~al\mbox{.}}{2021}]%
        {hsieh2021srvarm}
\bibfield{author}{\bibinfo{person}{Tsung-Yu Hsieh}, \bibinfo{person}{Yiwei
  Sun}, \bibinfo{person}{Xianfeng Tang}, \bibinfo{person}{Suhang Wang}, {and}
  \bibinfo{person}{Vasant~G Honavar}.} \bibinfo{year}{2021}\natexlab{}.
\newblock \showarticletitle{SrVARM: State Regularized Vector Autoregressive
  Model for Joint Learning of Hidden State Transitions and State-Dependent
  Inter-Variable Dependencies from Multi-variate Time Series}. In
  \bibinfo{booktitle}{\emph{Proceedings of the Web Conference 2021}}.
  \bibinfo{pages}{2270--2280}.
\newblock


\bibitem[\protect\citeauthoryear{Hussein, Gaber, Elyan, and Jayne}{Hussein
  et~al\mbox{.}}{2017}]%
        {hussein2017imitation}
\bibfield{author}{\bibinfo{person}{Ahmed Hussein},
  \bibinfo{person}{Mohamed~Medhat Gaber}, \bibinfo{person}{Eyad Elyan}, {and}
  \bibinfo{person}{Chrisina Jayne}.} \bibinfo{year}{2017}\natexlab{}.
\newblock \showarticletitle{Imitation learning: A survey of learning methods}.
\newblock \bibinfo{journal}{\emph{ACM Computing Surveys (CSUR)}}
  \bibinfo{volume}{50}, \bibinfo{number}{2} (\bibinfo{year}{2017}),
  \bibinfo{pages}{1--35}.
\newblock


\bibitem[\protect\citeauthoryear{Jin, Song, Li, Gai, Wang, and Zhang}{Jin
  et~al\mbox{.}}{2018}]%
        {jin2018real}
\bibfield{author}{\bibinfo{person}{Junqi Jin}, \bibinfo{person}{Chengru Song},
  \bibinfo{person}{Han Li}, \bibinfo{person}{Kun Gai}, \bibinfo{person}{Jun
  Wang}, {and} \bibinfo{person}{Weinan Zhang}.}
  \bibinfo{year}{2018}\natexlab{}.
\newblock \showarticletitle{Real-time bidding with multi-agent reinforcement
  learning in display advertising}. In \bibinfo{booktitle}{\emph{Proceedings of
  the 27th ACM International Conference on Information and Knowledge
  Management}}. \bibinfo{pages}{2193--2201}.
\newblock


\bibitem[\protect\citeauthoryear{Johnson, Bulgarelli, Pollard, Horng, Celi, and
  Mark~IV}{Johnson et~al\mbox{.}}{2020}]%
        {johnson2020mimic}
\bibfield{author}{\bibinfo{person}{Alistair Johnson}, \bibinfo{person}{Lucas
  Bulgarelli}, \bibinfo{person}{Tom Pollard}, \bibinfo{person}{Steven Horng},
  \bibinfo{person}{Leo~Anthony Celi}, {and} \bibinfo{person}{R Mark~IV}.}
  \bibinfo{year}{2020}\natexlab{}.
\newblock \showarticletitle{Mimic-iv (version 0.4)}.
\newblock \bibinfo{journal}{\emph{PhysioNet}} (\bibinfo{year}{2020}).
\newblock


\bibitem[\protect\citeauthoryear{Khanna and Tan}{Khanna and Tan}{2019}]%
        {khanna2019economy}
\bibfield{author}{\bibinfo{person}{Saurabh Khanna} {and}
  \bibinfo{person}{Vincent~YF Tan}.} \bibinfo{year}{2019}\natexlab{}.
\newblock \showarticletitle{Economy Statistical Recurrent Units For Inferring
  Nonlinear Granger Causality}. In \bibinfo{booktitle}{\emph{International
  Conference on Learning Representations}}.
\newblock


\bibitem[\protect\citeauthoryear{Kipf and Welling}{Kipf and Welling}{2016}]%
        {kipf2016semi}
\bibfield{author}{\bibinfo{person}{Thomas~N Kipf} {and} \bibinfo{person}{Max
  Welling}.} \bibinfo{year}{2016}\natexlab{}.
\newblock \showarticletitle{Semi-supervised classification with graph
  convolutional networks}.
\newblock \bibinfo{journal}{\emph{arXiv preprint arXiv:1609.02907}}
  (\bibinfo{year}{2016}).
\newblock


\bibitem[\protect\citeauthoryear{Kober and Peters}{Kober and Peters}{2009}]%
        {kober2009learning}
\bibfield{author}{\bibinfo{person}{Jens Kober} {and} \bibinfo{person}{Jan
  Peters}.} \bibinfo{year}{2009}\natexlab{}.
\newblock \showarticletitle{Learning motor primitives for robotics}. In
  \bibinfo{booktitle}{\emph{2009 IEEE International Conference on Robotics and
  Automation}}. IEEE, \bibinfo{pages}{2112--2118}.
\newblock


\bibitem[\protect\citeauthoryear{Li, Wang, Zhang, Chang, Liu, Xie, Qi, and
  Song}{Li et~al\mbox{.}}{2020}]%
        {li2020temporal}
\bibfield{author}{\bibinfo{person}{Shuang Li}, \bibinfo{person}{Lu Wang},
  \bibinfo{person}{Ruizhi Zhang}, \bibinfo{person}{Xiaofu Chang},
  \bibinfo{person}{Xuqin Liu}, \bibinfo{person}{Yao Xie}, \bibinfo{person}{Yuan
  Qi}, {and} \bibinfo{person}{Le Song}.} \bibinfo{year}{2020}\natexlab{}.
\newblock \showarticletitle{Temporal Logic Point Processes}. In
  \bibinfo{booktitle}{\emph{International Conference on Machine Learning}}.
  PMLR, \bibinfo{pages}{5990--6000}.
\newblock


\bibitem[\protect\citeauthoryear{Liang, Meng, Liu, Liu, Tu, Wang, Zhou, and
  Liu}{Liang et~al\mbox{.}}{2023a}]%
        {liang2023learn}
\bibfield{author}{\bibinfo{person}{Ke Liang}, \bibinfo{person}{Lingyuan Meng},
  \bibinfo{person}{Meng Liu}, \bibinfo{person}{Yue Liu},
  \bibinfo{person}{Wenxuan Tu}, \bibinfo{person}{Siwei Wang},
  \bibinfo{person}{Sihang Zhou}, {and} \bibinfo{person}{Xinwang Liu}.}
  \bibinfo{year}{2023}\natexlab{a}.
\newblock \showarticletitle{Learn from relational correlations and periodic
  events for temporal knowledge graph reasoning}. In
  \bibinfo{booktitle}{\emph{Proceedings of the 46th International ACM SIGIR
  Conference on Research and Development in Information Retrieval}}.
  \bibinfo{pages}{1559--1568}.
\newblock


\bibitem[\protect\citeauthoryear{Liang, Zhou, Liu, Meng, Liu, and Liu}{Liang
  et~al\mbox{.}}{2023b}]%
        {liang2023structure}
\bibfield{author}{\bibinfo{person}{Ke Liang}, \bibinfo{person}{Sihang Zhou},
  \bibinfo{person}{Yue Liu}, \bibinfo{person}{Lingyuan Meng},
  \bibinfo{person}{Meng Liu}, {and} \bibinfo{person}{Xinwang Liu}.}
  \bibinfo{year}{2023}\natexlab{b}.
\newblock \showarticletitle{Structure Guided Multi-modal Pre-trained
  Transformer for Knowledge Graph Reasoning}.
\newblock \bibinfo{journal}{\emph{arXiv preprint arXiv:2307.03591}}
  (\bibinfo{year}{2023}).
\newblock


\bibitem[\protect\citeauthoryear{L{\"o}we, Madras, Zemel, and Welling}{L{\"o}we
  et~al\mbox{.}}{2020}]%
        {lowe2020amortized}
\bibfield{author}{\bibinfo{person}{Sindy L{\"o}we}, \bibinfo{person}{David
  Madras}, \bibinfo{person}{Richard Zemel}, {and} \bibinfo{person}{Max
  Welling}.} \bibinfo{year}{2020}\natexlab{}.
\newblock \showarticletitle{Amortized causal discovery: Learning to infer
  causal graphs from time-series data}.
\newblock \bibinfo{journal}{\emph{arXiv preprint arXiv:2006.10833}}
  (\bibinfo{year}{2020}).
\newblock


\bibitem[\protect\citeauthoryear{Lyu, Yang, Liu, and Gustafson}{Lyu
  et~al\mbox{.}}{2019}]%
        {lyu2019sdrl}
\bibfield{author}{\bibinfo{person}{Daoming Lyu}, \bibinfo{person}{Fangkai
  Yang}, \bibinfo{person}{Bo Liu}, {and} \bibinfo{person}{Steven Gustafson}.}
  \bibinfo{year}{2019}\natexlab{}.
\newblock \showarticletitle{SDRL: interpretable and data-efficient deep
  reinforcement learning leveraging symbolic planning}. In
  \bibinfo{booktitle}{\emph{Proceedings of the AAAI Conference on Artificial
  Intelligence}}, Vol.~\bibinfo{volume}{33}. \bibinfo{pages}{2970--2977}.
\newblock


\bibitem[\protect\citeauthoryear{Meng, Han, Liu, and Tong}{Meng
  et~al\mbox{.}}{2018}]%
        {meng2018improving}
\bibfield{author}{\bibinfo{person}{Zibo Meng}, \bibinfo{person}{Shizhong Han},
  \bibinfo{person}{Ping Liu}, {and} \bibinfo{person}{Yan Tong}.}
  \bibinfo{year}{2018}\natexlab{}.
\newblock \showarticletitle{Improving speech related facial action unit
  recognition by audiovisual information fusion}.
\newblock \bibinfo{journal}{\emph{IEEE transactions on cybernetics}}
  \bibinfo{volume}{49}, \bibinfo{number}{9} (\bibinfo{year}{2018}),
  \bibinfo{pages}{3293--3306}.
\newblock


\bibitem[\protect\citeauthoryear{Mott, Zoran, Chrzanowski, Wierstra, and
  Jimenez~Rezende}{Mott et~al\mbox{.}}{2019}]%
        {mott2019towards}
\bibfield{author}{\bibinfo{person}{Alexander Mott}, \bibinfo{person}{Daniel
  Zoran}, \bibinfo{person}{Mike Chrzanowski}, \bibinfo{person}{Daan Wierstra},
  {and} \bibinfo{person}{Danilo Jimenez~Rezende}.}
  \bibinfo{year}{2019}\natexlab{}.
\newblock \showarticletitle{Towards Interpretable Reinforcement Learning Using
  Attention Augmented Agents}.
\newblock \bibinfo{journal}{\emph{Advances in Neural Information Processing
  Systems}}  \bibinfo{volume}{32} (\bibinfo{year}{2019}),
  \bibinfo{pages}{12350--12359}.
\newblock


\bibitem[\protect\citeauthoryear{Pamfil, Sriwattanaworachai, Desai,
  Pilgerstorfer, Georgatzis, Beaumont, and Aragam}{Pamfil
  et~al\mbox{.}}{2020}]%
        {pamfil2020dynotears}
\bibfield{author}{\bibinfo{person}{Roxana Pamfil}, \bibinfo{person}{Nisara
  Sriwattanaworachai}, \bibinfo{person}{Shaan Desai}, \bibinfo{person}{Philip
  Pilgerstorfer}, \bibinfo{person}{Konstantinos Georgatzis},
  \bibinfo{person}{Paul Beaumont}, {and} \bibinfo{person}{Bryon Aragam}.}
  \bibinfo{year}{2020}\natexlab{}.
\newblock \showarticletitle{Dynotears: Structure learning from time-series
  data}. In \bibinfo{booktitle}{\emph{International Conference on Artificial
  Intelligence and Statistics}}. PMLR, \bibinfo{pages}{1595--1605}.
\newblock


\bibitem[\protect\citeauthoryear{Qin, Chen, Wang, Lan, Ren, and Hong}{Qin
  et~al\mbox{.}}{2023}]%
        {qin2023read}
\bibfield{author}{\bibinfo{person}{Wei Qin}, \bibinfo{person}{Zetong Chen},
  \bibinfo{person}{Lei Wang}, \bibinfo{person}{Yunshi Lan},
  \bibinfo{person}{Weijieying Ren}, {and} \bibinfo{person}{Richang Hong}.}
  \bibinfo{year}{2023}\natexlab{}.
\newblock \showarticletitle{Read, Diagnose and Chat: Towards Explainable and
  Interactive LLMs-Augmented Depression Detection in Social Media}.
\newblock \bibinfo{journal}{\emph{arXiv preprint arXiv:2305.05138}}
  (\bibinfo{year}{2023}).
\newblock


\bibitem[\protect\citeauthoryear{Ren, Wang, Liu, Guo, Peng, and Fu}{Ren
  et~al\mbox{.}}{2022b}]%
        {ren2022mitigating}
\bibfield{author}{\bibinfo{person}{Weijieying Ren}, \bibinfo{person}{Lei Wang},
  \bibinfo{person}{Kunpeng Liu}, \bibinfo{person}{Ruocheng Guo},
  \bibinfo{person}{Lim~Ee Peng}, {and} \bibinfo{person}{Yanjie Fu}.}
  \bibinfo{year}{2022}\natexlab{b}.
\newblock \showarticletitle{Mitigating popularity bias in recommendation with
  unbalanced interactions: A gradient perspective}. In
  \bibinfo{booktitle}{\emph{2022 IEEE International Conference on Data Mining
  (ICDM)}}. IEEE, \bibinfo{pages}{438--447}.
\newblock


\bibitem[\protect\citeauthoryear{Ren, Wang, Li, Hughes, and Fu}{Ren
  et~al\mbox{.}}{2022a}]%
        {ren2022semi}
\bibfield{author}{\bibinfo{person}{Weijieying Ren}, \bibinfo{person}{Pengyang
  Wang}, \bibinfo{person}{Xiaolin Li}, \bibinfo{person}{Charles~E Hughes},
  {and} \bibinfo{person}{Yanjie Fu}.} \bibinfo{year}{2022}\natexlab{a}.
\newblock \showarticletitle{Semi-supervised drifted stream learning with short
  lookback}. In \bibinfo{booktitle}{\emph{Proceedings of the 28th ACM SIGKDD
  Conference on Knowledge Discovery and Data Mining}}.
  \bibinfo{pages}{1504--1513}.
\newblock


\bibitem[\protect\citeauthoryear{Ren, Zhang, Jiang, Wang, Guo, and Liu}{Ren
  et~al\mbox{.}}{2017}]%
        {ren2017robust}
\bibfield{author}{\bibinfo{person}{Weijieying Ren}, \bibinfo{person}{Lei
  Zhang}, \bibinfo{person}{Bo Jiang}, \bibinfo{person}{Zhefeng Wang},
  \bibinfo{person}{Guangming Guo}, {and} \bibinfo{person}{Guiquan Liu}.}
  \bibinfo{year}{2017}\natexlab{}.
\newblock \showarticletitle{Robust mapping learning for multi-view multi-label
  classification with missing labels}. In \bibinfo{booktitle}{\emph{Knowledge
  Science, Engineering and Management: 10th International Conference, KSEM
  2017, Melbourne, VIC, Australia, August 19-20, 2017, Proceedings 10}}.
  Springer, \bibinfo{pages}{543--551}.
\newblock


\bibitem[\protect\citeauthoryear{Ren, Jiang, Khoo, and Chieu}{Ren
  et~al\mbox{.}}{2021}]%
        {ren2021cross}
\bibfield{author}{\bibinfo{person}{Xiaoying Ren}, \bibinfo{person}{Jing Jiang},
  \bibinfo{person}{Ling Min~Serena Khoo}, {and} \bibinfo{person}{Hai~Leong
  Chieu}.} \bibinfo{year}{2021}\natexlab{}.
\newblock \showarticletitle{Cross-Topic Rumor Detection using Topic-Mixtures}.
  In \bibinfo{booktitle}{\emph{Proceedings of the 16th Conference of the
  European Chapter of the Association for Computational Linguistics: Main
  Volume}}. \bibinfo{pages}{1534--1538}.
\newblock


\bibitem[\protect\citeauthoryear{Tank, Covert, Foti, Shojaie, and Fox}{Tank
  et~al\mbox{.}}{2018}]%
        {tank2018neural}
\bibfield{author}{\bibinfo{person}{Alex Tank}, \bibinfo{person}{Ian Covert},
  \bibinfo{person}{Nicholas Foti}, \bibinfo{person}{Ali Shojaie}, {and}
  \bibinfo{person}{Emily Fox}.} \bibinfo{year}{2018}\natexlab{}.
\newblock \showarticletitle{Neural Granger Causality}.
\newblock \bibinfo{journal}{\emph{arXiv preprint arXiv:1802.05842}}
  (\bibinfo{year}{2018}).
\newblock


\bibitem[\protect\citeauthoryear{Tomasi, Tozzo, Salzo, and Verri}{Tomasi
  et~al\mbox{.}}{2018}]%
        {tomasi2018latent}
\bibfield{author}{\bibinfo{person}{Federico Tomasi}, \bibinfo{person}{Veronica
  Tozzo}, \bibinfo{person}{Saverio Salzo}, {and} \bibinfo{person}{Alessandro
  Verri}.} \bibinfo{year}{2018}\natexlab{}.
\newblock \showarticletitle{Latent variable time-varying network inference}. In
  \bibinfo{booktitle}{\emph{Proceedings of the 24th ACM SIGKDD International
  Conference on Knowledge Discovery \& Data Mining}}.
  \bibinfo{pages}{2338--2346}.
\newblock


\bibitem[\protect\citeauthoryear{Torabi, Warnell, and Stone}{Torabi
  et~al\mbox{.}}{2018}]%
        {torabi2018behavioral}
\bibfield{author}{\bibinfo{person}{Faraz Torabi}, \bibinfo{person}{Garrett
  Warnell}, {and} \bibinfo{person}{Peter Stone}.}
  \bibinfo{year}{2018}\natexlab{}.
\newblock \showarticletitle{Behavioral cloning from observation}. In
  \bibinfo{booktitle}{\emph{Proceedings of the 27th International Joint
  Conference on Artificial Intelligence}}. \bibinfo{pages}{4950--4957}.
\newblock


\bibitem[\protect\citeauthoryear{Van~Gerven, Taal, and Lucas}{Van~Gerven
  et~al\mbox{.}}{2008}]%
        {van2008dynamic}
\bibfield{author}{\bibinfo{person}{Marcel~AJ Van~Gerven},
  \bibinfo{person}{Babs~G Taal}, {and} \bibinfo{person}{Peter~JF Lucas}.}
  \bibinfo{year}{2008}\natexlab{}.
\newblock \showarticletitle{Dynamic Bayesian networks as prognostic models for
  clinical patient management}.
\newblock \bibinfo{journal}{\emph{Journal of biomedical informatics}}
  \bibinfo{volume}{41}, \bibinfo{number}{4} (\bibinfo{year}{2008}),
  \bibinfo{pages}{515--529}.
\newblock


\bibitem[\protect\citeauthoryear{Wang, Yu, He, Cheng, Ren, Wang, Zong, Chen,
  and Zha}{Wang et~al\mbox{.}}{2020}]%
        {wang2020adversarial}
\bibfield{author}{\bibinfo{person}{Lu Wang}, \bibinfo{person}{Wenchao Yu},
  \bibinfo{person}{Xiaofeng He}, \bibinfo{person}{Wei Cheng},
  \bibinfo{person}{Martin~Renqiang Ren}, \bibinfo{person}{Wei Wang},
  \bibinfo{person}{Bo Zong}, \bibinfo{person}{Haifeng Chen}, {and}
  \bibinfo{person}{Hongyuan Zha}.} \bibinfo{year}{2020}\natexlab{}.
\newblock \showarticletitle{Adversarial Cooperative Imitation Learning for
  Dynamic Treatment Regimes}. In \bibinfo{booktitle}{\emph{Proceedings of The
  Web Conference 2020}}. \bibinfo{pages}{1785--1795}.
\newblock


\bibitem[\protect\citeauthoryear{Widrow}{Widrow}{1964}]%
        {widrow1964pattern}
\bibfield{author}{\bibinfo{person}{Bernard Widrow}.}
  \bibinfo{year}{1964}\natexlab{}.
\newblock \showarticletitle{Pattern-recognizing control systems}.
\newblock \bibinfo{journal}{\emph{Compurter and Information Sciences}}
  (\bibinfo{year}{1964}).
\newblock


\bibitem[\protect\citeauthoryear{Yu, Chen, Gao, and Yu}{Yu
  et~al\mbox{.}}{2019}]%
        {yu2019dag}
\bibfield{author}{\bibinfo{person}{Yue Yu}, \bibinfo{person}{Jie Chen},
  \bibinfo{person}{Tian Gao}, {and} \bibinfo{person}{Mo Yu}.}
  \bibinfo{year}{2019}\natexlab{}.
\newblock \showarticletitle{Dag-gnn: Dag structure learning with graph neural
  networks}. In \bibinfo{booktitle}{\emph{International Conference on Machine
  Learning}}. PMLR, \bibinfo{pages}{7154--7163}.
\newblock


\bibitem[\protect\citeauthoryear{Zahavy, Ben-Zrihem, and Mannor}{Zahavy
  et~al\mbox{.}}{2016}]%
        {zahavy2016graying}
\bibfield{author}{\bibinfo{person}{Tom Zahavy}, \bibinfo{person}{Nir
  Ben-Zrihem}, {and} \bibinfo{person}{Shie Mannor}.}
  \bibinfo{year}{2016}\natexlab{}.
\newblock \showarticletitle{Graying the black box: Understanding dqns}. In
  \bibinfo{booktitle}{\emph{International Conference on Machine Learning}}.
  PMLR, \bibinfo{pages}{1899--1908}.
\newblock


\bibitem[\protect\citeauthoryear{Zambaldi, Raposo, Santoro, Bapst, Li,
  Babuschkin, Tuyls, Reichert, Lillicrap, Lockhart, et~al\mbox{.}}{Zambaldi
  et~al\mbox{.}}{2018}]%
        {zambaldi2018relational}
\bibfield{author}{\bibinfo{person}{Vinicius Zambaldi}, \bibinfo{person}{David
  Raposo}, \bibinfo{person}{Adam Santoro}, \bibinfo{person}{Victor Bapst},
  \bibinfo{person}{Yujia Li}, \bibinfo{person}{Igor Babuschkin},
  \bibinfo{person}{Karl Tuyls}, \bibinfo{person}{David Reichert},
  \bibinfo{person}{Timothy Lillicrap}, \bibinfo{person}{Edward Lockhart},
  {et~al\mbox{.}}} \bibinfo{year}{2018}\natexlab{}.
\newblock \showarticletitle{Relational deep reinforcement learning}.
\newblock \bibinfo{journal}{\emph{arXiv preprint arXiv:1806.01830}}
  (\bibinfo{year}{2018}).
\newblock


\bibitem[\protect\citeauthoryear{Zhao, Liu, Huang, Li, Liu, GuiQuan, and
  Shi}{Zhao et~al\mbox{.}}{2020}]%
        {zhao2020balancing}
\bibfield{author}{\bibinfo{person}{Tianxiang Zhao}, \bibinfo{person}{Lemao
  Liu}, \bibinfo{person}{Guoping Huang}, \bibinfo{person}{Huayang Li},
  \bibinfo{person}{Yingling Liu}, \bibinfo{person}{Liu GuiQuan}, {and}
  \bibinfo{person}{Shuming Shi}.} \bibinfo{year}{2020}\natexlab{}.
\newblock \showarticletitle{Balancing quality and human involvement: An
  effective approach to interactive neural machine translation}. In
  \bibinfo{booktitle}{\emph{Proceedings of the AAAI Conference on Artificial
  Intelligence}}, Vol.~\bibinfo{volume}{34}. \bibinfo{pages}{9660--9667}.
\newblock


\bibitem[\protect\citeauthoryear{Zhao, Luo, Zhang, and Wang}{Zhao
  et~al\mbox{.}}{2023a}]%
        {zhao2023towards}
\bibfield{author}{\bibinfo{person}{Tianxiang Zhao}, \bibinfo{person}{Dongsheng
  Luo}, \bibinfo{person}{Xiang Zhang}, {and} \bibinfo{person}{Suhang Wang}.}
  \bibinfo{year}{2023}\natexlab{a}.
\newblock \showarticletitle{Towards Faithful and Consistent Explanations for
  Graph Neural Networks}. In \bibinfo{booktitle}{\emph{Proceedings of the
  Sixteenth ACM International Conference on Web Search and Data Mining}}.
  \bibinfo{pages}{634--642}.
\newblock


\bibitem[\protect\citeauthoryear{Zhao, Yu, Wang, Wang, Zhang, Chen, Liu, Cheng,
  and Chen}{Zhao et~al\mbox{.}}{2023b}]%
        {zhao2023skill}
\bibfield{author}{\bibinfo{person}{Tianxiang Zhao}, \bibinfo{person}{Wenchao
  Yu}, \bibinfo{person}{Suhang Wang}, \bibinfo{person}{Lu Wang},
  \bibinfo{person}{Xiang Zhang}, \bibinfo{person}{Yuncong Chen},
  \bibinfo{person}{Yanchi Liu}, \bibinfo{person}{Wei Cheng}, {and}
  \bibinfo{person}{Haifeng Chen}.} \bibinfo{year}{2023}\natexlab{b}.
\newblock \showarticletitle{Skill Disentanglement for Imitation Learning from
  Suboptimal Demonstrations}.
\newblock \bibinfo{journal}{\emph{arXiv preprint arXiv:2306.07919}}
  (\bibinfo{year}{2023}).
\newblock


\bibitem[\protect\citeauthoryear{Zheng, Aragam, Ravikumar, and Xing}{Zheng
  et~al\mbox{.}}{2018}]%
        {zheng2018dags}
\bibfield{author}{\bibinfo{person}{Xun Zheng}, \bibinfo{person}{Bryon Aragam},
  \bibinfo{person}{Pradeep~K Ravikumar}, {and} \bibinfo{person}{Eric~P Xing}.}
  \bibinfo{year}{2018}\natexlab{}.
\newblock \showarticletitle{DAGs with NO TEARS: Continuous Optimization for
  Structure Learning}.
\newblock \bibinfo{journal}{\emph{Advances in Neural Information Processing
  Systems}}  \bibinfo{volume}{31} (\bibinfo{year}{2018}).
\newblock


\bibitem[\protect\citeauthoryear{Zheng, Das, Young, Swanson, Warren, and
  Sarkar}{Zheng et~al\mbox{.}}{2014}]%
        {zheng2014autonomous}
\bibfield{author}{\bibinfo{person}{Zhi Zheng}, \bibinfo{person}{Shuvajit Das},
  \bibinfo{person}{Eric~M Young}, \bibinfo{person}{Amy Swanson},
  \bibinfo{person}{Zachary Warren}, {and} \bibinfo{person}{Nilanjan Sarkar}.}
  \bibinfo{year}{2014}\natexlab{}.
\newblock \showarticletitle{Autonomous robot-mediated imitation learning for
  children with autism}. In \bibinfo{booktitle}{\emph{2014 IEEE International
  Conference on Robotics and Automation (ICRA)}}. IEEE,
  \bibinfo{pages}{2707--2712}.
\newblock


\bibitem[\protect\citeauthoryear{Ziebart, Maas, Bagnell, Dey,
  et~al\mbox{.}}{Ziebart et~al\mbox{.}}{2008}]%
        {ziebart2008maximum}
\bibfield{author}{\bibinfo{person}{Brian~D Ziebart}, \bibinfo{person}{Andrew~L
  Maas}, \bibinfo{person}{J~Andrew Bagnell}, \bibinfo{person}{Anind~K Dey},
  {et~al\mbox{.}}} \bibinfo{year}{2008}\natexlab{}.
\newblock \showarticletitle{Maximum entropy inverse reinforcement learning.}.
  In \bibinfo{booktitle}{\emph{Aaai}}, Vol.~\bibinfo{volume}{8}. Chicago, IL,
  USA, \bibinfo{pages}{1433--1438}.
\newblock


\end{thebibliography}
